\documentclass[11pt]{article}

\usepackage[preprint]{acl}

\usepackage{times}
\usepackage{latexsym}

\usepackage[T1]{fontenc}

\usepackage[utf8]{inputenc}

\usepackage{microtype}

\usepackage{inconsolata}

\usepackage{graphicx}

\usepackage{array}
\usepackage{longtable}
\usepackage{ragged2e}
\usepackage{caption}
\usepackage{wrapfig}
\usepackage{tabularx}
\usepackage{amsmath}
\usepackage{xspace}
\usepackage{cleveref}
\usepackage{placeins}
\usepackage{makecell}
\usepackage{multirow}

\usepackage{hyperref}       
\usepackage{url}            
\usepackage{booktabs}       
\usepackage{amsfonts}       
\usepackage{nicefrac}       
\usepackage[table]{xcolor}  

\definecolor{headerbg}{HTML}{DCE6F1}
\definecolor{headerline}{HTML}{5B6B7A}

\definecolor{rowblue}{HTML}{F4F8FC}
\definecolor{rowindigo}{HTML}{F5F3FA}
\definecolor{rowrose}{HTML}{FCF4F6}
\definecolor{rowamber}{HTML}{FFF8EE}
\definecolor{rowgreen}{HTML}{F2F8F2}
\definecolor{rowteal}{HTML}{F1F8F8}

\newcolumntype{C}[1]{>{\centering\arraybackslash}p{#1}}

\usepackage[most]{tcolorbox}

\definecolor{keyboxbg}{HTML}{F7F9FC}
\definecolor{keyboxframe}{HTML}{5B6B7A}

\newtcolorbox{keybox}[1][]{
  enhanced,
  sharp corners,
  boxsep=0pt,
  left=4pt, right=4pt, top=4pt, bottom=4pt,
  before skip=3pt, after skip=3pt,
  colback=keyboxbg,
  colframe=keyboxframe,
  boxrule=.45pt,
  drop shadow={black!10},
  #1
}

\usepackage{dblfloatfix}

\newcommand{\webIcon}{\raisebox{-2pt}{\includegraphics[height=1.15em]{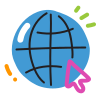}}\xspace}

\newcommand{\github}{\raisebox{-1.5pt}{\includegraphics[height=1.05em]{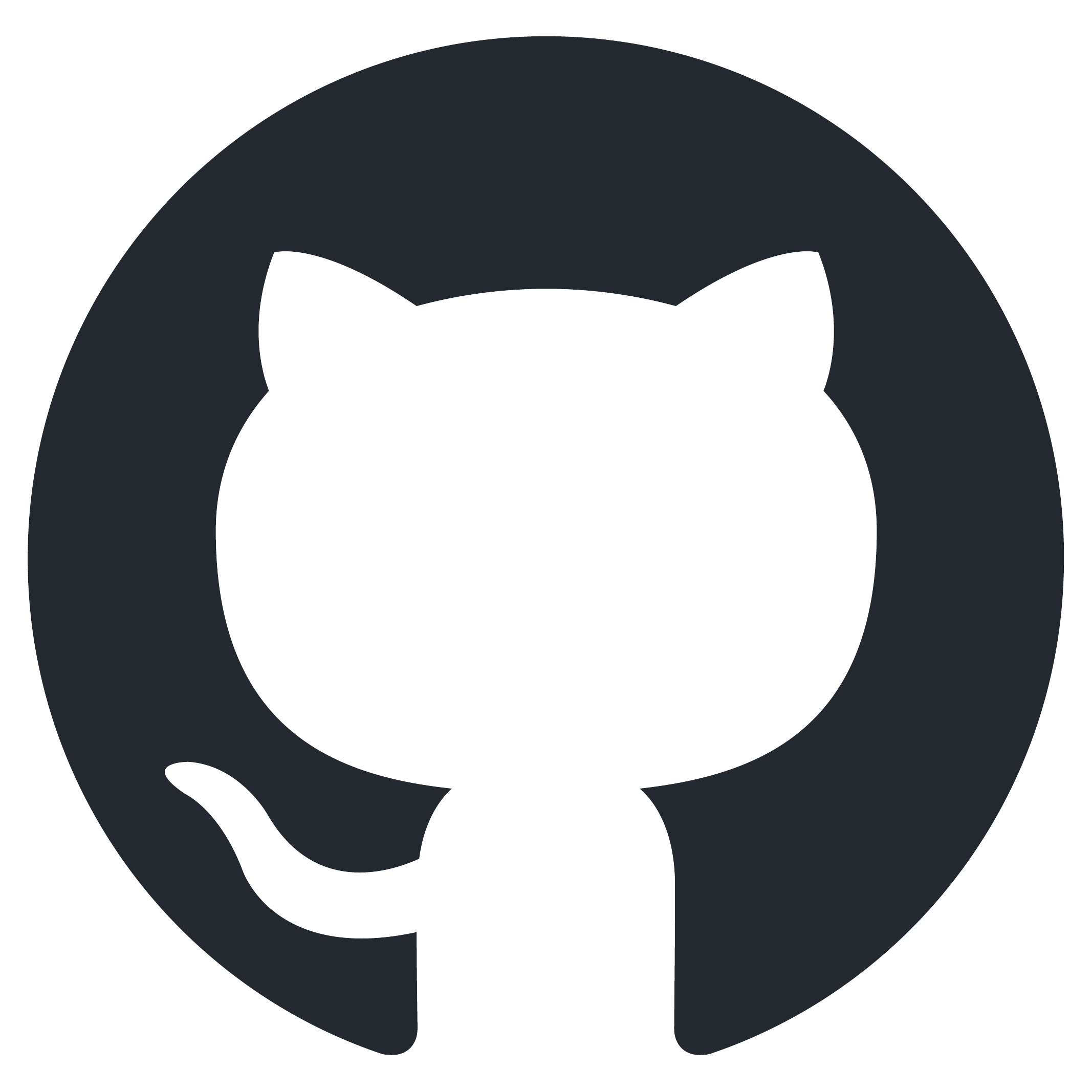}}\xspace}

\usepackage{tikz}
\usetikzlibrary{arrows.meta, positioning, shapes.multipart, fit}

\definecolor{headergray}{HTML}{E6E6E6}
\definecolor{rowgray}{HTML}{F7F7F7}
\newcommand{\LangRowStretch}{1.40}

\newcommand{\sref}[1]{\S\ref{#1}}

\title{Whose Data, Whose Benchmark?\\ A User-Centric Framework for Auditing Evaluation Resources}

\author{Mir Tafseer Nayeem \\
  University of Alberta \\
  \texttt{mnayeem@ualberta.ca} \\\And
  Davood Rafiei \\
  University of Alberta \\
  \texttt{drafiei@ualberta.ca} \\}

\begin{document}
\maketitle

\begin{abstract}
Benchmarks shape how model capabilities are measured and interpreted, yet their scores can obscure whose conventions, contexts, and judgments define successful performance. We introduce a \emph{user-centric audit framework} that distinguishes users who contribute benchmark content from those whose needs evaluation results are intended to represent. The framework operationalizes seven dimensions across benchmark content, metadata, documentation, and scoring procedures. Across 16 resources, we find that American-centered conventions account for 62.4--94.4\% of matched signals in 11 text benchmarks, while geographic visibility in five image--text resources is concentrated and strongly associated with national GDP share (mean Pearson $r=0.780$; Spearman $\rho=0.778$). Documentation reveals a systematic source--target gap: all 16 resources operationalize source-user context, but none operationalize target-user context. Paired scoring experiments further show that meaning-preserving answer variants reduce DROP exact-match accuracy from 100.0\% to 33.3\%, while TruthfulQA exhibits metric-dependent reductions in reference similarity. Our framework and documentation checklist make user assumptions auditable, clarifying what benchmark scores represent and where evaluation claims are supported.\footnote{\webIcon{} \github{} Data, code, and research artifacts are publicly available at \href{https://tafseer-nayeem.github.io/WhoseBench/}{\texttt{\path{tafseer-nayeem.github.io/WhoseBench}}}.}

\end{abstract}


\section{Introduction}
\label{sec:intro}

Large Language Models (LLMs) and multimodal AI systems are increasingly used across education, healthcare, information access, and everyday decision-making. Their users differ in language, cultural conventions, expertise, geographic context, and expectations. As these systems reach broader populations, benchmarks play a growing role in defining how capabilities are measured, compared, and communicated. Recent research has examined benchmark validity, evaluation reliability, and dataset documentation~\citep{raji2021ai,liu-etal-2024-ecbd,mousavi-etal-2026-garbage}. Yet a fundamental question remains: \emph{whose users are represented in the resources that define successful performance?}

Language and multimodal data are produced by people, interpreted by people, and deployed for people. Evaluation resources consequently inherit assumptions about the people and contexts from which their content originates. A benchmark may favor particular spelling, date, or measurement conventions; image--text collections may disproportionately reference certain countries; and accepted answers may depend on annotator judgments or scoring rules. Prior work has highlighted the importance of documenting data provenance and linguistic diversity~\citep{bender-friedman-2018-data,10.1145/3458723,faisal-etal-2024-dialectbench}. However, understanding how these assumptions collectively shape the scope and interpretation of benchmark claims remains an important challenge.

We distinguish between \emph{source users}, whose content and judgments contribute to evaluation resources, and \emph{target users}, whose needs and contexts the resulting scores are intended to inform. Documenting where benchmark content originates does not establish which user populations or settings its scores represent. This distinction motivates a \emph{user-centric audit}: a systematic examination of the assumptions embedded in evaluation resources and their implications for what benchmark scores measure and where evaluation claims apply. Such an audit considers whose conventions are treated as defaults, which geographies are visible, whose judgments define accepted answers, and whether intended evaluation contexts are documented.

We introduce a \emph{user-centric framework for auditing evaluation resources} that operationalizes these questions across \emph{seven} dimensions: linguistic and locale conventions, geographic visibility, socioeconomic structure, source-user context, target-user assumptions, annotator visibility, and evaluation normalization. By connecting evidence from benchmark content, metadata, documentation, and scoring procedures, the framework provides a unified approach to examining resource-level assumptions and assessing the scope of evaluation claims.

We instantiate the framework through three empirical audits spanning 16 resources. First, we examine American-centered linguistic and locale conventions across 11 text benchmarks, including how meaning-preserving answer variants affect scoring outcomes. Second, we measure country visibility in captions and associated metadata from five image--text resources and examine its relationship with socioeconomic structure. Third, we assess whether documentation across all 16 resources makes source users, target contexts, annotator information, locale assumptions, and evaluation procedures explicit and operationally inspectable. These complementary analyses connect benchmark composition, evaluation rules, and documented user contexts within a common framework.

Our contributions are threefold:
\begin{enumerate}
    \item We introduce a reusable \emph{user-centric audit framework} that distinguishes source users from target users and operationalizes seven dimensions of user-relevant assumptions across evaluation resources~(\sref{sec:user_centric_lens}).

    \item We systematically audit 16 text and image--text resources, revealing American-centered convention asymmetries~(\sref{sec:rq1_convention_audit}), geographic concentration associated with socioeconomic structure~(\sref{sec:rq2_geo_multimodal}), and a persistent source--target documentation gap~(\sref{sec:documentation_audit}). Paired scoring experiments further demonstrate how valid answer conventions can affect evaluation outcomes.

    \item We provide an actionable documentation checklist and reproducible audit artifacts that support benchmark creators and users in examining resource assumptions and interpreting the scope of evaluation claims~(\sref{sec:discussion}).
\end{enumerate}


\section{Related Work}
\label{sec:related_work}

\paragraph{Benchmark validity and documentation.}
Prior research establishes how benchmark design and construction shape the measurement and interpretation of model capabilities~\citep{PAULLADA2021100336,raji2021ai,Miller2022BenchmarkValidity,liu-etal-2024-ecbd}, including the influence of benchmark content and scoring procedures on evaluation reliability~\citep{mousavi-etal-2026-garbage}. Documentation frameworks, including data statements, datasheets, and BenchmarkCards, provide mechanisms for reporting provenance, intended use, and evaluation conditions~\citep{bender-friedman-2018-data,10.1145/3458723,sokol2025benchmarkcards}. We operationalize these concerns through a user-centric audit framework that connects resource composition, scoring conventions, and documented user contexts to the validity and scope of benchmark evaluation claims.

\paragraph{User-centric and linguistic evaluation.}
Research on user-centric evaluation addresses real-world user intents~\citep{wang-etal-2024-user-centric}, performance across dialects and language varieties~\citep{faisal-etal-2024-dialectbench}, cultural variation in time expressions~\citep{shwartz-2022-good}, and American English preferences across the LLM pipeline~\citep{nayeem2026englishllms}. 
Population-specific data construction for children's language~\citep{nayeem-rafiei-2024-kidlm} and personalized opinion summarization with context-sensitive evaluation~\citep{nayeem2025opiniorag} further demonstrate the importance of aligning resources and evaluation procedures with intended users. We extend these insights to the \emph{evaluation resources} themselves, systematically measuring linguistic and locale conventions, testing how scoring procedures handle meaning-preserving alternatives, and examining whether documented target-user assumptions support broader evaluation claims.

\paragraph{Geographic representation in multimodal resources.}
Prior studies identify uneven geographic coverage in visual datasets and regional disparities in model performance~\citep{shankar2017geodiversity,devries2019objectrecognition}. More recently, \citet{basu2026imagescomefrom} analyzed country representation in image--text datasets through caption-based geographic profiling. We integrate geographic visibility and its socioeconomic associations into the user-centric audit, connecting patterns of country representation with source provenance, intended user contexts, and the interpretation of evaluation results.

\begin{table*}[t]
\centering
\footnotesize
\renewcommand{\arraystretch}{1.3}
\setlength{\tabcolsep}{4.5pt}

\rowcolors{2}{rowgray}{white}
\resizebox{0.95\linewidth}{!}{%
\begin{tabularx}{\textwidth}{@{}p{0.21\textwidth} p{0.32\textwidth} X@{}}
\rowcolor{headergray}
\toprule
\textbf{Audit dimension} & \textbf{User-centric question} & \textbf{Operational signals in this paper} \\
\midrule

Linguistic and locale conventions
& Whose spelling, formatting, and measurement conventions are treated as default?
& Spelling variants (e.g., color/colour), date formats, time notation (e.g., 7 pm/19:00), and measurement units (e.g., miles/kilometres) in benchmark questions, passages, answer options, and gold answers.
 \\

Geographic visibility
& Which countries and regions are visible in image--text resource descriptions?
& Explicit country references in captions and textual metadata, resolved using country names, demonyms (e.g., Japanese), capitals, and selected cities (e.g., Paris). \\

Socioeconomic structure
& How does geographic visibility relate to economic and socioeconomic indicators?
& Associations with GDP share, population share, GDP per capita, Internet usage, and tourism arrivals. \\

Source-user context
& Where does benchmark content originate, and whose contexts shape its collection?
& Documented sources (e.g., examinations, Wikipedia, Reddit, CommonCrawl) and provenance metadata, such as website domains (e.g., \texttt{bbc.co.uk}, \texttt{reddit.com}) and country-code TLDs (e.g., \texttt{.uk}, \texttt{.ca}, \texttt{.jp}). \\

Target-user assumptions
& Which populations and settings are evaluation claims intended to represent?
& Documented intended users (e.g., students, clinicians), evaluation audiences, deployment contexts, and conditions for interpreting benchmark results. \\

Annotator visibility
& Whose judgments determine labels, references, and annotation quality?
& Documented annotator expertise (e.g., medical professionals), demographics, locale (e.g., country, language variety), recruitment (e.g., crowdsourcing), compensation, annotation protocols, and quality-control procedures (e.g., inter-annotator agreement). \\

Evaluation normalization
& Do scoring procedures recognize meaning-preserving answer alternatives?
& Documented normalization rules, accepted variants (e.g., color/colour), locale conventions, and paired scorer-sensitivity tests using valid answer alternatives. \\

\bottomrule
\end{tabularx}%
}
\vspace{-1.5mm}
\caption{\small User-centric audit taxonomy for evaluation resources. Seven dimensions connect user-relevant assumptions to measurable and inspectable evidence across benchmark content, metadata, documentation, and scoring procedures.}
\vspace{-2.75mm}
\label{tab:user_centric_audit_taxonomy}
\end{table*}


\section{A User-Centric Framework for Auditing Evaluation Resources}
\label{sec:user_centric_lens}

A user-centric audit begins with a premise: \emph{language and multimodal data are produced by people, interpreted by people, and deployed for people}. Evaluation resources therefore reflect choices about whose language, knowledge, experiences, and judgments are represented. Our framework systematically examines these choices, making the assumptions embedded in benchmark content, metadata, documentation, and scoring procedures explicit and auditable. This provides a basis for understanding what benchmark scores represent and the user contexts in which their interpretation is supported.

\paragraph{Source users and target users.}
We distinguish \emph{source users}, whose content and judgments contribute to an evaluation resource, from \emph{target users}, whose needs and contexts its results are intended to inform. Source users include exam writers, crowdworkers, and web contributors, while target users may include students, clinicians, and multilingual speakers. Consider a medical benchmark constructed from professional examination questions and subsequently used to evaluate systems intended for clinicians across healthcare settings. Its documentation may establish where the questions originated without specifying how the resulting scores apply across those settings. The source--target distinction thus connects benchmark provenance to the scope of its evaluation claims.

\begin{figure*}[t]
\centering
\includegraphics[width=0.94\linewidth]{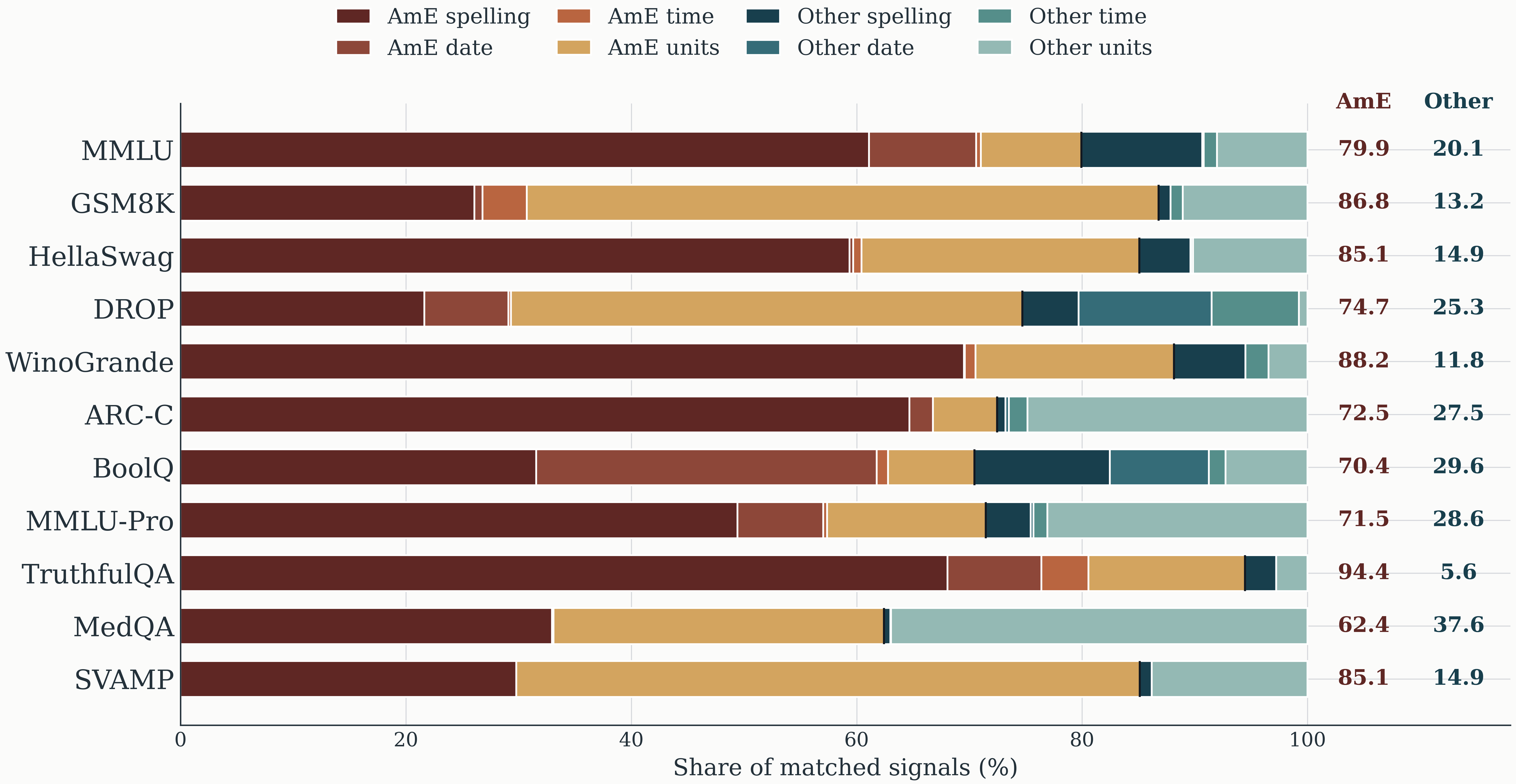}
\vspace{-1.0mm}
\caption{\small Linguistic and locale convention signals across 11 text benchmarks. Bars compare American-centered and other user-valid matches in spelling, dates, time, and measurement units; rightmost columns show aggregate shares. American-centered conventions predominate, with variation across benchmarks and convention types. Appendix Figure~\ref{fig:AmE_match_detail} provides the breakdown.
}
\label{fig:AmE_match_by_type}
\end{figure*}

\paragraph{Dimensions of a user-centric audit.}
Table~\ref{tab:user_centric_audit_taxonomy} organizes the framework into \emph{seven} dimensions: linguistic and locale conventions, geographic visibility, socioeconomic structure, source-user context, target-user assumptions, annotator visibility, and evaluation normalization. Each dimension translates a question about users into measurable or inspectable evidence drawn from benchmark content, metadata, documentation, or evaluation procedures. The resulting framework connects how a resource is constructed, whose contexts it represents, and how its scores are determined.

\paragraph{From resource assumptions to evaluation claims.}
Benchmark composition and scoring rules shape the evidence used to assess model capabilities. A meaning-preserving answer expressed as \textit{colour}, \textit{kilometres}, or a day--month date may receive a different score when the evaluator recognizes only another convention. Likewise, concentrated geographic references in image--text captions and metadata make some contexts more visible than others. These examples illustrate how linguistic defaults, geographic representation, and evaluation normalization influence the interpretation of benchmark results. Auditing these properties makes the underlying assumptions available for systematic examination before evaluation scores are generalized across users and settings.

We apply the framework through three complementary empirical audits: linguistic and locale conventions in text benchmarks~(\sref{sec:rq1_convention_audit}), geographic visibility and socioeconomic structure in image--text resources~(\sref{sec:rq2_geo_multimodal}), and documentation of user-relevant assumptions across all resources~(\sref{sec:documentation_audit}). A paired scoring-sensitivity analysis additionally tests how evaluation procedures respond to meaning-preserving answer variants~(\sref{sec:rq1_scorer_case_study}). This integrated approach connects the assumptions present in evaluation resources to the evidence and scoring decisions underlying benchmark claims.


\section{RQ1: Do Text Benchmarks Encode an Implicit Default User?}
\label{sec:rq1_convention_audit}

Text benchmarks assess knowledge, reasoning, and comprehension through questions and reference answers that also encode linguistic and locale conventions. We audit which conventions these resources favor and how valid differences in answer expression affect evaluation scores.

\paragraph{Benchmarks and audit setup.}
We examine 11 benchmarks spanning knowledge, mathematics, commonsense reasoning, reading comprehension, science, truthfulness, and medical QA: MMLU, GSM8K, HellaSwag, DROP, WinoGrande, ARC-Challenge, BoolQ, MMLU-Pro, TruthfulQA, MedQA, and SVAMP. Several feature prominently in model evaluations~\citep{grattafiori2024llama3,gemmateam2025gemma3}. Appendix Table~\ref{tab:benchmark-overview} summarizes their tasks and formats.

We audit four convention classes: spelling, date formats, time notation, and measurement units. English varieties have developed through distinct historical and sociolinguistic contexts, including the global spread of English and postcolonial language practices~\citep{crystal2003english,0b15e89a04f846adb87cf3cb75ed6a2b}. Some conventions are shared across regions, while others may differ. We group selected alternatives under \emph{other user-valid conventions} for comparative measurement, recognizing that these forms do not constitute a single English variety. Appendix~\ref{sec:appendix_audit_method} provides the convention inventory and detection rules.

\paragraph{Method.}
We detect American-centered and other user-valid convention signals in benchmark questions, passages, instructions, answer options, and gold answers. Each item is classified as American-centered only, other conventions only, mixed, or unmatched. We report \emph{coverage}, the percentage of items containing at least one audited signal, and \emph{American-centered share}, the percentage of matched signals belonging to the American-centered category. We also exclude spelling matches to assess whether the asymmetry extends to date, time, and measurement conventions. Appendix~\ref{sec:appendix_audit_method} details the aggregation procedure.

\subsection{Results and Analysis}
\label{sec:rq1_results_analysis}

\paragraph{American-centered conventions dominate matched signals.}
Figure~\ref{fig:AmE_match_by_type} shows an American-centered majority in all 11 benchmarks, ranging from 62.4\% in MedQA to 94.4\% in TruthfulQA. The pattern spans general knowledge, mathematics, commonsense reasoning, and specialized QA.

Coverage varies substantially across resources. Table~\ref{tab:rq1_coverage_robustness} shows that audited conventions occur in 5.97\% of WinoGrande items and 7.59\% of TruthfulQA items, compared with 65.85\% in MedQA and 78.08\% in DROP. American-centered share captures the distribution of \emph{matched signals}, while coverage measures their prevalence across benchmark items. Reporting both distinguishes convention prevalence from which forms predominate.

\begin{table}[t]
\centering
\small
\renewcommand{\arraystretch}{1.15}
\setlength{\tabcolsep}{5pt}

\resizebox{0.95\columnwidth}{!}{%
\begin{tabular}{@{}lrr@{}}
\toprule
\textbf{Benchmark}
& \textbf{Coverage (\%)}
& \shortstack{\textbf{AmE excl.}\\\textbf{spelling (\%)}} \\
\midrule

MMLU~\citep{hendrycks2021mmlu}
& 26.60 & 66.9 \\

GSM8K~\citep{cobbe2021gsm8k}
& 25.40 & 83.3 \\

HellaSwag~\citep{zellers2019hellaswag}
& 38.11 & 71.2 \\

DROP~\citep{dua2019drop}
& 78.08 & 72.3 \\

WinoGrande~\citep{sakaguchi2020winogrande}
& 5.97 & 77.2 \\

ARC-Challenge~\citep{clark2018arc}
& 18.73 & 22.5 \\

BoolQ~\citep{clark2019boolq}
& 44.92 & 68.9 \\

MMLU-Pro~\citep{wang2024mmlupro}
& 37.80 & 47.3 \\

TruthfulQA~\citep{lin2022truthfulqa}
& 7.59 & 90.5 \\

MedQA~\citep{jin2021medqa}
& 65.85 & 44.3 \\

SVAMP~\citep{patel2021svamp}
& 9.30 & 80.0 \\

\bottomrule
\end{tabular}%
}
\vspace{-6pt}
\caption{\small Convention coverage and spelling-excluded robustness across 11 text benchmarks. Coverage is the share of items containing an audited signal; the final column reports the American-centered share among non-spelling matches.
\vspace{-6pt}
}
\label{tab:rq1_coverage_robustness}
\end{table}

\paragraph{The asymmetry extends beyond spelling.}
Excluding spelling matches leaves an American-centered majority in eight of 11 benchmarks. The highest shares occur in TruthfulQA (90.5\%), GSM8K (83.3\%), SVAMP (80.0\%), and DROP (72.3\%). ARC-Challenge (22.5\%), MedQA (44.3\%), and MMLU-Pro (47.3\%) show the opposite pattern. These results reveal convention profiles across benchmarks and establish that the asymmetry extends beyond orthographic variation.

\subsection{Scorer Case Study: Do Valid Answer Variants Change Evaluation Scores?}
\label{sec:rq1_scorer_case_study}

The convention audit identifies which forms appear in benchmark content. We next test whether \emph{meaning-preserving changes in answer conventions alter evaluation outcomes} using controlled answer pairs and scoring procedures. This resource-level analysis isolates scoring sensitivity from \emph{model-specific} behavior, establishing how evaluation rules recognize valid answer alternatives.

\paragraph{Paired scoring experiment.}
We select DROP and TruthfulQA for their released answer-text evaluation procedures. For DROP, we use the official exact-match (EM) and token-level F1 scorer~\citep{dua2019drop}; for TruthfulQA, we reproduce the released BLEU/ROUGE correct-versus-incorrect reference-ranking procedure~\citep{lin2022truthfulqa}. Other benchmarks, such as MMLU and ARC-Challenge, primarily score multiple-choice labels, while GSM8K evaluates extracted numerical answers. Prior research on multiple-choice evaluation and free-form answer matching further motivates directly testing how scoring procedures handle valid answer alternatives~\citep{zheng2024mcselectors,wang-etal-2024-answer-c,chandak2025answermatching}.

\begin{table}[t]
\centering
\footnotesize
\renewcommand{\arraystretch}{1.16}
\setlength{\tabcolsep}{3.5pt}

\resizebox{0.95\columnwidth}{!}{%
\begin{tabular}{@{}llrrr@{}}
\toprule
\textbf{Benchmark} & \textbf{Metric}
& \textbf{Orig.} & \textbf{Var.}
& \textbf{$\Delta$ [95\% CI]} \\
\midrule
DROP & EM
& 100.0 & 33.3
& $-66.7$ [$-92.3,-36.4$] \\
DROP & F1
& 100.0 & 77.2
& $-22.8$ [$-41.6,-5.2$] \\
\midrule
TQA & BLEU correct.
& 97.0 & 97.0
& $0.0$ [$0.0,0.0$] \\
TQA & ROUGE-1 correct.
& 97.0 & 97.0
& $0.0$ [$0.0,0.0$] \\
TQA & ROUGE-2 correct.
& 100.0 & 93.9
& $-6.1$ [$-16.1,0.0$] \\
TQA & ROUGE-1 sim.
& 100.0 & 88.9
& $-11.1$ [$-13.5,-9.0$] \\
\bottomrule
\end{tabular}%
}
\vspace{-3pt}
\caption{\small Scorer sensitivity to answer variants. DROP uses official EM/F1; TQA (TruthfulQA) uses BLEU/ROUGE reference-ranking rules. Changes are in percentage points for EM, F1, and correctness, and on a 0--100 scale for similarity. CIs use question-clustered bootstrap resampling.
\vspace{-4pt}
}
\label{tab:answer_surface_sensitivity}
\end{table}

For each eligible pair, we compare an original reference-consistent answer with a meaning-preserving spelling, written-date, or time-format variant against the same gold annotations. We retain each evaluation procedure and estimate paired changes with 95\% confidence intervals using question-clustered bootstrapping. Appendix~\ref{app:rq1_scoring_details} details the transformations, screening, sample sizes, scoring details, and uncertainty estimates.

\paragraph{DROP: Valid alternatives can lose substantial credit.}
Table~\ref{tab:answer_surface_sensitivity} shows that DROP's official EM decreases from 100.0\% to 33.3\% for meaning-preserving variants, a reduction of 66.7 percentage points (95\% CI $[-92.3,-36.4]$). Token-level F1 also decreases from 100.0\% to 77.2\% ($-22.8$ points; 95\% CI $[-41.6,-5.2]$). Partial credit reduces the penalty, but valid changes in answer expression still affect evaluation scores.

\paragraph{TruthfulQA: Lower similarity need not change correctness.}
TruthfulQA responds differently. BLEU and ROUGE-1 correctness remain at 97.0\%, even as ROUGE-1 similarity to a correct reference decreases from 100.0 to 88.9 (95\% CI for the change $[-13.5,-9.0]$). ROUGE-2 correctness declines by 6.1 points, with a confidence interval that includes zero. Valid variants can reduce reference similarity while preserving the evaluator's correctness decision. The contrast with DROP demonstrates how scoring rules determine the consequences of differences in answer expression.

\begin{figure*}[t]
\centering
\includegraphics[width=0.97\textwidth]{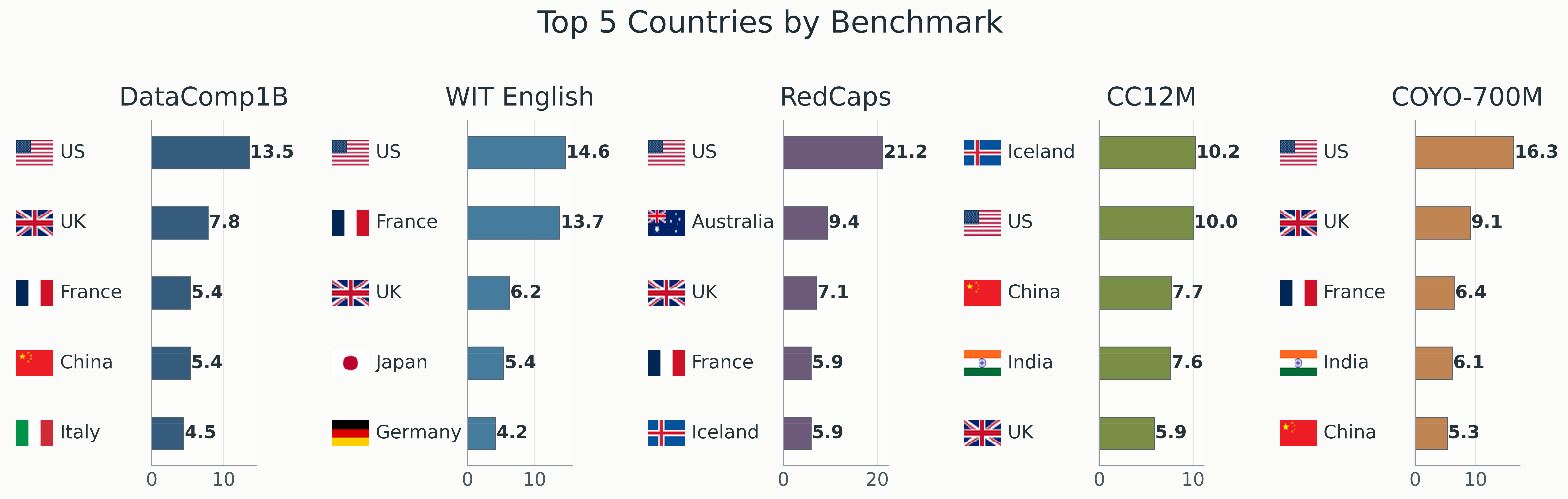}
\vspace{-1.0mm}
\caption{\textbf{Caption geography is concentrated in a few countries.} Bars show the top five country shares from caption-derived assignments in each resource. The US leads four resources, while Iceland leads CC12M, revealing source-specific variation. Countries are identified through deterministic text matching, without image geolocation.
}
\label{fig:top5_by_benchmark}
\end{figure*}

\begin{figure*}[t]
\centering
\includegraphics[width=0.97\textwidth]{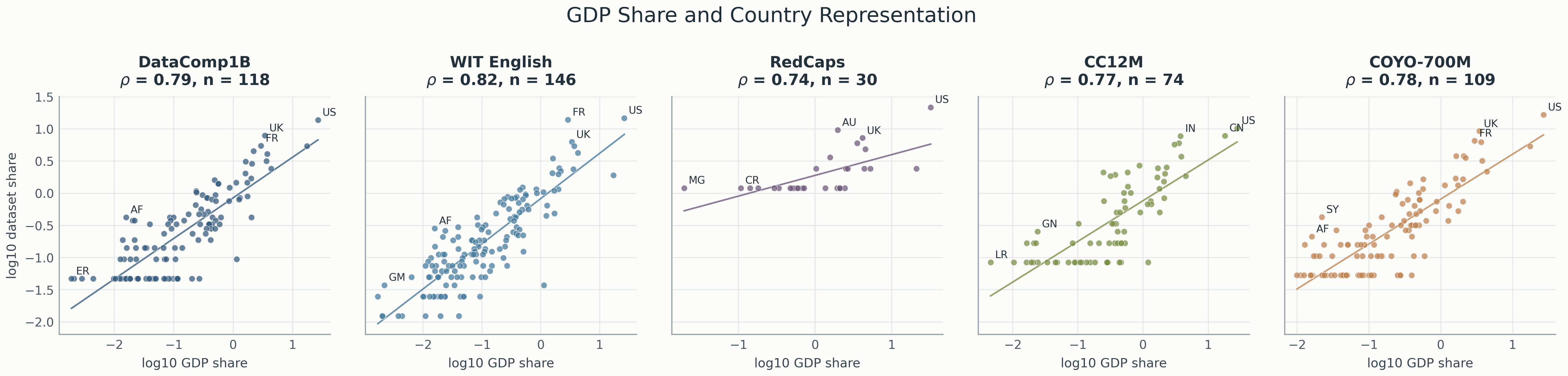}
\vspace{-1.0mm}
\caption{\textbf{Country visibility is strongly associated with GDP share.} Each point represents a country with at least one million people. GDP and caption-derived country shares are normalized over the same retained countries and plotted on log scales. Fitted trends and Spearman correlations summarize the association across resources.
}
\label{fig:gdp_share_vs_representation}
\end{figure*}

\paragraph{Implications for user-aware evaluation.}
These findings distinguish \emph{conventions represented in benchmark content} from \emph{alternatives recognized by scoring procedures}. Exact match, token overlap, and reference ranking can score meaning-preserving responses differently. Benchmark creators can test consistency across spelling, date, time, and measurement variants and document normalization rules. Such audits reveal how answer conventions influence evaluation outcomes and inform the user-centric documentation checklist~(\sref{sec:discussion}).


\section{RQ2: Which Geographies Are Visible in Multimodal Benchmarks?}
\label{sec:rq2_geo_multimodal}

Image--text resources reflect geographic contexts through captions and associated metadata, shaping which countries become explicitly visible in multimodal data. We audit the distribution of these geographic references across five resources and examine their relationship with national socioeconomic indicators and language composition.

\paragraph{Resources and audit setup.}
We analyze five image--text resources with distinct collection sources: DataComp1B, WIT English, RedCaps, CC12M, and COYO-700M. These span web-scale image--text collections, Wikipedia metadata, Reddit captions, web alt-text, and CommonCrawl-derived pairs, respectively (Appendix Table~\ref{tab:image-text-benchmarks}).

We identify explicit geographic references in captions and associated textual metadata using a deterministic resolver covering country names, aliases, demonyms, capitals, and selected cities. Records are assigned to countries when the textual evidence yields a unique resolution. We define \emph{geographic visibility} as the distribution of country assignments among resolved records, measuring which geographies appear explicitly in each resource's textual layer. Appendix~\ref{app:caption_geo_experiments} details the lexicon, matching rules, and assignment procedure.

\begin{table*}[!t]
\centering
\small
\renewcommand{\arraystretch}{1.16}
\setlength{\tabcolsep}{2pt}
\resizebox{0.98\linewidth}{!}{%
\begin{tabularx}{\textwidth}{@{}p{0.22\textwidth} C{0.08\textwidth} X@{}}
\toprule
\rowcolor{headerbg}
\textbf{Criterion}
& \textbf{Score}
& \textbf{Audit question} \\
\midrule

\rowcolor{rowblue}
Source-user context
& 0--2
& Does the documentation identify who produced the content, including source populations, platforms (e.g., Reddit, Wikipedia), institutions, or collection processes?
 \\

\rowcolor{rowindigo}
Target-user or evaluation context
& 0--2
& Does it specify intended users (e.g., students, clinicians), evaluation audiences, deployment settings, or the contexts in which benchmark results are expected to apply? \\

\rowcolor{rowrose}
Annotator visibility
& 0--2
& Does it describe annotator expertise, demographics, locale, recruitment (e.g., crowdsourcing), compensation, quality control, or the absence of human annotation? \\

\rowcolor{rowamber}
Locale assumptions
& 0--2
& Does it document spelling (e.g., color/colour), date, time, measurement, currency, numeric-format, language-variety, or source-language conventions? \\

\rowcolor{rowgreen}
Evaluation normalization
& 0--2
& Does it document how scoring procedures handle valid alternative answers, formatting conventions (e.g., date formats), or locale-specific equivalents? \\

\rowcolor{rowteal}
Source/geographic metadata
& 0--2
& Does it expose source URLs, website domains (e.g., \texttt{bbc.co.uk}, \texttt{reddit.com}), top-level domains (e.g., \texttt{.uk}, \texttt{.ca}, \texttt{.jp}), language tags (e.g., \texttt{en}, \texttt{fr}, \texttt{ja}), country or region fields, or other metadata supporting provenance and geographic auditing?
 \\

\bottomrule
\end{tabularx}
}
\vspace{-3pt}
\caption{\small Documentation audit rubric across six user-relevant criteria. Each criterion is scored as 0 (absent), 1 (mentioned explicitly), or 2 (operationalized through inspectable data, metadata, or evaluation procedures).
}
\label{tab:documentation_audit_rubric}
\end{table*}

\subsection{Results and Analysis}
\label{sec:rq2_results_analysis}

\paragraph{Geographic visibility is concentrated across sources.}
Figure~\ref{fig:top5_by_benchmark} shows that the United States leads four resources: DataComp1B (13.55\%), WIT English (14.59\%), RedCaps (21.18\%), and COYO-700M (16.30\%). CC12M differs, with Iceland first~(10.24\%) and the United States second~(9.99\%). Across resources, the top three countries account for 26.8--37.6\% of resolved assignments, while Europe, North America, and Oceania collectively account for 60.9--76.5\%. These patterns reveal geographic concentration across distinct collection sources. Appendix~\ref{app:top_bottom_country_analyses} reports the top-ten distributions and low-frequency countries.

\paragraph{Country visibility closely tracks economic scale.}
Figure~\ref{fig:gdp_share_vs_representation} shows a strong association between country visibility and World Bank GDP share (mean Pearson $r=0.780$; Spearman $\rho=0.778$), with rank correlations ranging from $\rho=0.735$ in RedCaps to $\rho=0.823$ in WIT English. GDP share provides the strongest single-variable fit among socioeconomic indicators (mean $R^2=0.614$; count-weighted $R^2=0.748$). Population share, GDP per capita, Internet usage, and tourism arrivals yield weaker fits and modest improvements when combined with GDP share. These associations are descriptive rather than causal. Appendix~\ref{app:gdp_control_variables} details the regression specifications and comparisons.

\paragraph{Language composition changes visibility, but economic concentration persists.}
To assess language effects, we apply the same resolver to a balanced WIT sample spanning French, German, Spanish, and Japanese, motivated by documented concentration in English webtext~\citep{dodge-etal-2021-documenting}. In this separate comparison, the U.S. share falls from 20.0\% in English to 8.5\% in the non-English sample. Yet the ten largest economies account for 53.5\% of resolved assignments, with strong GDP associations (Pearson $r=0.752$; Spearman $\rho=0.744$). Language composition shifts geographic visibility, while economic concentration persists~(see Appendix~\ref{app:wit_language_stress_test} for details).

\paragraph{Implications for user-aware evaluation.}
Geographic visibility varies with resource composition and language and is strongly associated with economic scale. Country references identify contexts visible in resource text without establishing the intended evaluation users. Documenting geographic coverage, languages, and source contexts makes these distinctions auditable and informs the user-centric documentation checklist~(\sref{sec:discussion}).


\begin{table}[t]
\centering
\small
\renewcommand{\arraystretch}{1.14}
\setlength{\tabcolsep}{3.2pt}
\resizebox{0.92\linewidth}{!}{%
\begin{tabularx}{\columnwidth}{
    >{\raggedright\arraybackslash}X
    >{\centering\arraybackslash}p{0.13\columnwidth}
    >{\centering\arraybackslash}p{0.13\columnwidth}
    >{\centering\arraybackslash}p{0.13\columnwidth}
}
\toprule
\rowcolor{headergray}
\textbf{Criterion}
& \makecell{\textbf{Mean}\\\textbf{score}}
& \makecell{\textbf{Ment.}\\\textbf{(\%)}}
& \makecell{\textbf{Oper.}\\\textbf{(\%)}} \\
\midrule

Source-user context       & 2.00 & 100.0 & 100.0 \\
Target/eval. context      & 1.00 & 100.0 & 0.0   \\
Annotator visibility      & 0.56 & 56.2  & 0.0   \\
Locale assumptions        & 0.31 & 25.0  & 6.2   \\
Evaluation normalization  & 1.44 & 75.0  & 68.8  \\
Source/geographic metadata & 0.69 & 37.5 & 31.2 \\

\bottomrule
\end{tabularx}
}
\vspace{-2.5pt}
\caption{\small Documentation Audit results across 16 resources. Mean scores range from 0 (absent) to 2 (operationalized). ``Ment.'' indicates resources scoring 1 or 2; ``Oper.'' indicates those scoring 2, \% are computed across all 16 resources.
\vspace{-3.5pt}
}
\label{tab:documentation_audit_results}
\end{table}

\section{RQ3: Are User Assumptions Made Explicit in Documentation?}
\label{sec:documentation_audit}

Benchmark documentation defines how evaluation resources can be understood, inspected, and applied. Building on the linguistic and geographic audits in \sref{sec:rq1_convention_audit} and \sref{sec:rq2_geo_multimodal}, we assess whether documentation makes six user-relevant dimensions explicit: source-user context, target-user context, annotator visibility, locale assumptions, evaluation normalization, and source/geographic metadata.

\begin{table*}[t]
\centering
\footnotesize
\renewcommand{\arraystretch}{1.16}
\setlength{\tabcolsep}{4pt}
\rowcolors{2}{rowgray}{white}

\resizebox{0.96\linewidth}{!}{%
\begin{tabularx}{\textwidth}{
@{}p{0.19\textwidth}
   p{0.40\textwidth}
   X@{}
}
\toprule
\rowcolor{headergray}
\textbf{Field}
& \textbf{Minimum documentation question}
& \textbf{Why it matters for evaluation} \\
\midrule

Source-user context
& Who produced the content, and through which platform, institution, task, or collection process?
& Identifies whose language, knowledge, experiences, and judgments are represented. \\

Target-user context
& Which users, settings, or evaluation claims is the benchmark intended to represent?
& Clarifies the populations and contexts to which benchmark results may apply. \\

Annotator context
& Who provided labels or reference answers, and what is known about their expertise, locale, recruitment, compensation, and quality control?
& Makes visible whose judgments determine correctness, relevance, or acceptability. \\

Locale conventions
& Which spelling, date, time, unit, currency, numeric-format, or language-variety conventions appear in the data or scoring procedure?
& Identifies regional assumptions that may influence benchmark content and evaluation outcomes. \\

Evaluation normalization
& How does the scorer handle valid alternative answers, including spelling, formatting, measurement, and locale-specific equivalents?
& Reveals whether equivalent responses receive consistent scores across user conventions. \\

Source/geographic metadata
& Which URLs, website domains, top-level domains, language tags, or geographic fields are available for auditing?
& Supports tracing content sources and examining geographic visibility without conflating source provenance with content location. \\

\bottomrule
\end{tabularx}
}
\vspace{-3.5pt}
\caption{\small Minimal user-centric documentation checklist for evaluation resources. The six fields translate the audit taxonomy into practical reporting questions that clarify benchmark provenance, intended users, and evaluation assumptions.
\vspace{-6.5pt}
}
\label{tab:user_centric_documentation_checklist}
\end{table*}

\paragraph{Resources and audit setup.}
We audit public documentation for all 16 resources: 11 text benchmarks and five image--text collections (Appendix Tables~\ref{tab:benchmark-overview} and~\ref{tab:image-text-benchmarks}). Sources include benchmark papers, official dataset cards, and repositories, where available (Appendix Table~\ref{tab:documentation_audit_sources}).

\paragraph{Scoring protocol.}
Two annotators independently score each resource against the six criteria in Table~\ref{tab:documentation_audit_rubric}: \textbf{0} (absent), \textbf{1} (mentioned), or \textbf{2} (operationalized). A score of 1 requires an explicit statement; a score of 2 requires an inspectable mechanism, such as metadata fields, source partitions, annotation protocols, or evaluation rules. Each resource receives the highest score supported by documented evidence. Appendix~\ref{sec:documentation_audit_agreement} reports the annotation procedure and agreement.

\subsection{Results and Analysis}
\label{sec:documentation_audit_results}

Table~\ref{tab:documentation_audit_results} reports average scores and proportions of resources mentioning or operationalizing criteria.

\paragraph{Documentation identifies sources more clearly than intended users.}
All 16 resources operationalize source-user context through documented origins, including examinations, Wikipedia, Reddit, web alt-text, and CommonCrawl. Although all mention target-user or evaluation context, none operationalize it. This systematic gap shows that resource provenance is more clearly specified than the populations, settings, and needs that benchmark scores are intended to represent.

\paragraph{Annotators and locale conventions remain under-documented.}
Annotator information is mentioned in 56.2\% of resources, with none operationalizing annotator attributes as auditable metadata. Locale assumptions receive even less attention: 25.0\% mention them, and only 6.2\% operationalize them. Consequently, the human judgments and linguistic conventions underlying evaluation references remain difficult to inspect.

\paragraph{Scoring rules are clearer than audit-enabling metadata.}
Evaluation normalization is mentioned in 75.0\% of resources and operationalized in 68.8\%. However, documented scoring procedures may still treat meaning-preserving answer variants differently, as demonstrated in \sref{sec:rq1_scorer_case_study}. Source/geographic metadata is mentioned in 37.5\% of resources and operationalized in 31.2\%. URLs and Internet domains support source tracing, while country-code domains (e.g., \texttt{.uk}, \texttt{.ca}) offer geographic clues without directly identifying content origin or contributor location. These gaps motivate the user-centric documentation checklist~(\sref{sec:discussion}).


\section{Discussion: Toward User-Centric Benchmark Design}
\label{sec:discussion}

Our three audits show that evaluation resources encode user-relevant assumptions through content, metadata, documentation, and scoring procedures. Linguistic conventions affect which answers actually receive credit, geographic references reveal uneven country visibility, and documentation specifies data provenance more clearly than intended evaluation users. These findings connect resource design to benchmark validity: interpreting scores requires understanding whose conventions, contexts, and judgments define successful performance and where evaluation claims reliably apply.

A user-centric approach makes these assumptions \emph{explicit} and \emph{inspectable}. Benchmark creators should explicitly document whose data and judgments shape a resource, which populations and settings its evaluation claims represent, and how scoring procedures handle valid linguistic and locale variations. Table~\ref{tab:user_centric_documentation_checklist} translates these requirements into \emph{six practical documentation fields}, with an extended reporting template supporting implementation in Appendix~\ref{app:user_centric_template}.

\paragraph{Implications for benchmark design and use.}
For creators, the checklist supports clearly specifying intended users, documenting scoring conventions, and identifying validation needs. For benchmark users, it helps assess whether reported scores apply to specific populations, languages, locales, and deployment settings. Making these assumptions inspectable also enables researchers to critically examine unsupported generalizations and assess the scope of evaluation claims.

\vspace{0.75mm}
\begin{keybox}
\small \textbf{Auditing evaluation begins with the resource.}
Benchmark content, metadata, reference answers, and scoring procedures shape evaluation before models are evaluated. Making their assumptions explicit strengthens score interpretation and clarifies the scope of claims across user contexts.
\end{keybox}


\section{Conclusion}
\label{sec:conclusion}

We introduced a user-centric framework for auditing evaluation resources and applied it to 11 text benchmarks and five image--text resources. Our findings reveal American-centered linguistic conventions, geographic visibility associated with national GDP share, and gaps between documented data sources and intended users. Paired scoring experiments show that meaning-preserving answer variants can receive different scores. These findings highlight how assumptions shape evaluation and why making them explicit is essential for interpreting benchmark scores across user contexts.



\section*{Limitations}
\label{sec:limitations}

\paragraph{Linguistic conventions and scoring.}
Our linguistic audit examines explicit spelling, date, time, and measurement conventions, providing a reproducible account of observable locale assumptions. Paired experiments using the official DROP scorer and reproduced TruthfulQA generation evaluator quantify how meaning-preserving answer variants affect evaluation outcomes. These analyses characterize linguistic representation and scoring sensitivity within the audited resources.

\paragraph{Geographic visibility and socioeconomic structure.}
We measure geographic visibility through explicit country references in captions and associated textual metadata, using deterministic resolution to support consistent cross-resource comparisons. Country shares reflect resolved textual references, while socioeconomic correlations characterize associations between geographic visibility and national economic indicators.

\paragraph{Public documentation and auditability.}
Our documentation audit examines publicly available benchmark papers, dataset cards, and repositories. The rubric assesses whether user-relevant assumptions are explicitly documented and operationalized through inspectable evidence. Its scores therefore characterize the transparency and auditability of released documentation.

\paragraph{Empirical scope and extensibility.}
Our study examines 11 text benchmarks and five image--text resources, covering selected linguistic, geographic, and socioeconomic dimensions. The framework provides a foundation for broader audits spanning multilingual evaluation, accessibility, diverse user contexts, and richer annotation metadata.


\section*{Ethics Statement}
\label{sec:ethics}

\paragraph{Data use and privacy.}
This study audits publicly available evaluation resources, documentation, textual metadata, and socioeconomic indicators. Analyses use research materials and aggregate measurements, respecting dataset licenses and usage conditions. Geographic visibility is measured through captions and associated metadata, with country assignments representing textual references rather than image provenance or contributor locations.

\paragraph{Responsible interpretation.}
Our findings characterize linguistic and locale conventions, geographic visibility, socioeconomic associations, scoring sensitivity, and documentation practices. These measurements describe resource-level properties and evaluation procedures, supporting interpretation of benchmark assumptions without inferring individual identities or demographic characteristics. Socioeconomic relationships are reported as descriptive associations.

\paragraph{User-centric accountability.}
Our framework encourages documentation of data provenance, intended users, annotator practices, and scoring conventions. These disclosures help benchmark creators define the scope of evaluation claims and enable researchers to assess their applicability across linguistic, geographic, and user contexts.

\paragraph{Open science and reproducibility.}
We publicly release code and research artifacts, including convention detectors, country-resolution rules, aggregate results, regression analyses, and documentation audit materials.\footnote{\webIcon{} \github{} Data, code, and research artifacts are available at \href{https://tafseer-nayeem.github.io/WhoseBench/}{\texttt{\path{tafseer-nayeem.github.io/WhoseBench}}}.} Our code is licensed under Apache 2.0, while original research artifacts are released under Creative Commons Attribution-NonCommercial-ShareAlike 4.0 International (CC BY-NC-SA 4.0).\footnote{\url{https://creativecommons.org/licenses/by-nc-sa/4.0/}} These materials support reproducibility and future research efforts while respecting the applicable licenses and distribution restrictions of the original resources.


\bibliography{custom}


\clearpage
\appendix

\twocolumn[{%
\centering
{\Large\bfseries Supplementary Material: Appendices}

\vspace{9pt}

\begin{tcolorbox}[
    enhanced,
    colback=gray!2,
    colframe=gray!28,
    boxrule=0.4pt,
    arc=2pt,
    left=9pt,
    right=9pt,
    top=7pt,
    bottom=7pt,
    width=0.94\textwidth
]

\centering
{\large\bfseries Appendix Contents}

\vspace{6pt}

\begin{minipage}{0.90\textwidth}
\raggedright
\small

\noindent
\hyperref[sec:appendix_audit_method]{%
\textbf{A\quad Benchmark Convention Audit}}

\vspace{2pt}

\hspace*{1.6em}
\hyperref[sec:appendix_benchmark_selection]{%
A.1\quad Benchmark Selection and Evaluation Text}\\

\hspace*{1.6em}
\hyperref[sec:appendix_variation_classes]{%
A.2\quad Convention Inventory and Detection Rules}\\

\hspace*{1.6em}
\hyperref[sec:appendix_item_labeling]{%
A.3\quad Item Labels and Aggregation Metrics}\\

\hspace*{1.6em}
\hyperref[sec:appendix_convention_distributions]{%
A.4\quad Convention-Level Distributions}\\

\hspace*{1.6em}
\hyperref[sec:appendix_coverage_robustness]{%
A.5\quad Coverage and Spelling-Excluded Robustness}\\

\hspace*{1.6em}
\hyperref[app:rq1_scoring_details]{%
A.6\quad Scorer Case Study: Construction and Method}\\

\hspace*{1.6em}
\hyperref[sec:appendix_scoring_results]{%
A.7\quad Scorer Case Study: Detailed Results}

\vspace{4pt}

\noindent
\hyperref[app:caption_geo_experiments]{%
\textbf{B\quad Geographic Visibility Audit}}

\vspace{2pt}

\hspace*{1.6em}
\hyperref[app:benchmarks]{%
B.1\quad Image--Text Resources}\\

\hspace*{1.6em}
\hyperref[app:country_lexicon_construction]{%
B.2\quad Country Lexicon Construction}\\

\hspace*{1.6em}
\hyperref[app:matching_and_country_assignment]{%
B.3\quad Matching and Country Assignment}\\

\hspace*{1.6em}
\hyperref[app:top_bottom_country_analyses]{%
B.4\quad Geographic Concentration}\\

\hspace*{1.6em}
\hyperref[app:gdp_correlation_experiment]{%
B.5\quad GDP and Country Representation}\\

\hspace*{1.6em}
\hyperref[app:gdp_control_variables]{%
B.6\quad Socioeconomic Comparisons and Controls}\\

\hspace*{1.6em}
\hyperref[app:wit_language_stress_test]{%
B.7\quad Language-Composition Stress Test on WIT}

\vspace{4pt}

\noindent
\hyperref[sec:documentation_audit_agreement]{%
\textbf{C\quad Documentation Audit}}

\vspace{4pt}

\noindent
\hyperref[app:user_centric_template]{%
\textbf{D\quad User-Centric Documentation Template}}

\vspace{4pt}

\noindent
\hyperref[sec:AI-usage]{%
\textbf{E\quad AI Use Statement}}

\end{minipage}

\end{tcolorbox}

\vspace{7pt}
}]


\section{Benchmark Convention Audit}
\label{sec:appendix_audit_method}

This appendix documents the RQ1 convention audit, covering benchmark selection, detection rules, aggregation metrics, spelling-excluded robustness, and the paired scorer study (\cref{sec:rq1_convention_audit}).

\subsection{Benchmark Selection and Evaluation Text}
\label{sec:appendix_benchmark_selection}

We audit 11 text benchmarks: MMLU, GSM8K, HellaSwag, DROP, WinoGrande, ARC-Challenge, BoolQ, MMLU-Pro, TruthfulQA, MedQA, and SVAMP. These resources span professional knowledge, mathematical reasoning, commonsense reasoning, reading comprehension, science, truthfulness, medical question answering, and robustness-oriented mathematical reasoning. Table~\ref{tab:benchmark-overview} summarizes their tasks and evaluation formats.

Each benchmark item is the unit of analysis. We concatenate questions, passages, instructions, answer options, and gold answers to capture the linguistic conventions presented to readers and used in evaluation. The audit characterizes these resource texts independently of model-generated responses.

\begin{table*}[!tp]
\centering
\renewcommand{\arraystretch}{\LangRowStretch}
\setlength{\tabcolsep}{3.2pt}
\rowcolors{2}{rowgray}{white}
\begin{tabular}{
  >{\RaggedRight\arraybackslash}p{0.18\textwidth}
  >{\RaggedRight\arraybackslash}p{0.20\textwidth}
  >{\RaggedRight\arraybackslash}p{0.18\textwidth}
  >{\RaggedRight\arraybackslash}p{0.38\textwidth}
}
\rowcolor{headergray}
\toprule
\textbf{Benchmark} &
\textbf{Task} &
\textbf{Category} &
\textbf{Short description} \\
\midrule

MMLU~\citep{hendrycks2021mmlu}
& Multiple-choice QA over 57 subjects
& Broad knowledge and reasoning
& Academic and professional exam questions spanning humanities, STEM, and professional domains. \\

GSM8K~\citep{cobbe2021gsm8k}
& Grade-school math word problems
& Mathematical reasoning
& Linguistically diverse elementary math problems with free-form reasoning and numeric answers. \\

HellaSwag~\citep{zellers2019hellaswag}
& Sentence / ending completion
& Commonsense reasoning
& Adversarially filtered everyday-event completion benchmark with multiple-choice endings. \\

DROP~\citep{dua2019drop}
& Reading comprehension with arithmetic
& Reading + discrete reasoning
& Paragraph-based QA requiring counting, comparison, addition, subtraction, and date/number tracking. \\

WinoGrande~\citep{sakaguchi2020winogrande}
& Pronoun / referent resolution
& Commonsense coreference
& Large-scale adversarial Winograd-style benchmark for commonsense pronoun resolution. \\

ARC-Challenge~\citep{clark2018arc}
& Science multiple-choice QA
& Science reasoning
& Difficult subset of grade-school science questions from the AI2 Reasoning Challenge. \\

BoolQ~\citep{clark2019boolq}
& Yes/no passage QA
& Factual QA
& Naturally occurring yes/no questions paired with short passages, originally sourced from search queries. \\

MMLU-Pro~\citep{wang2024mmlupro}
& Harder multiple-choice QA
& Broad knowledge + reasoning
& Harder successor to MMLU with more reasoning-focused items and expanded answer-option sets. \\

TruthfulQA~\citep{lin2022truthfulqa}
& Open-ended and multiple-choice QA
& Truthfulness / factuality
& Questions designed to elicit common misconceptions and imitative falsehoods. \\

MedQA~\citep{jin2021medqa}
& Medical exam multiple-choice QA
& Domain-specific professional QA
& Medical board-style exam questions requiring specialized clinical knowledge and decision-making. \\

SVAMP~\citep{patel2021svamp}
& Perturbed math word problems
& Robustness in math reasoning
& Challenge set built by applying controlled variations to simple arithmetic word problems. \\

\bottomrule
\end{tabular}
\caption{Overview of the benchmarks used in our evaluation, summarizing each benchmark’s task formulation, capability category, and intended reasoning or knowledge focus.}
\label{tab:benchmark-overview}
\end{table*}

\subsection{Convention Inventory and Detection Rules}
\label{sec:appendix_variation_classes}

We examine four classes of explicit linguistic and locale
conventions:

\begin{enumerate}
    \item \textbf{Spelling.} Orthographic alternatives,
    including \textit{color}/\textit{colour} and
    \textit{center}/\textit{centre}.

    \item \textbf{Date formats.} Numeric and written
    date conventions, including
    \texttt{04/21/2025}/\texttt{21/04/2025} and
    \textit{April 21, 2025}/\textit{21 April 2025}.

    \item \textbf{Time formats.} Twelve-hour and
    twenty-four-hour expressions, such as
    \textit{6:30 pm}/\texttt{18:30}.

    \item \textbf{Measurement units.} Contrasts involving
    U.S.-customary and metric conventions, including
    \textit{miles}/\textit{kilometres},
    \textit{pounds}/\textit{kilograms}, and
    \textit{Fahrenheit}/\textit{Celsius}.
\end{enumerate}

Spelling detection uses the curated orthographic inventory of \citet{nayeem2026englishllms}. Pattern-based detectors identify date, time, and measurement conventions in benchmark text, drawing on temporal distinctions discussed by \citet{shwartz-2022-good}.

Detection targets explicit surface forms. Spelling variants require lexical matches, preventing partial matches inside unrelated words. Date and time expressions are identified by structural patterns; abbreviated units additionally require quantitative context. These rules provide consistent, inspectable evidence across benchmarks.

The American-centered category comprises the audited American English spellings, month--day dates, twelve-hour times, and U.S.-customary measurement signals. The comparison category contains the corresponding other user-valid forms. The labels describe the conventions, independently of the identity or nationality of item contributors. Table~\ref{tab:example_contrasts} illustrates these convention contrasts.

\begin{table*}[!tp]
\centering
\small
\renewcommand{\arraystretch}{1.18}
\setlength{\tabcolsep}{6pt}

\begin{tabularx}{\textwidth}{@{}lXXl@{}}
\toprule
\textbf{Convention}
& \textbf{American-centered example}
& \textbf{Other user-valid example}
& \textbf{Signal type} \\
\midrule

Spelling
& \textit{color}, \textit{center}
& \textit{colour}, \textit{centre}
& Orthographic \\

Numeric date
& \texttt{04/21/2025}
& \texttt{21/04/2025}
& Date order \\

Written date
& \textit{April 21, 2025}
& \textit{21 April 2025}
& Date order \\

Time notation
& \textit{7:00 pm}
& \texttt{19:00}
& Time format \\

Measurement
& \textit{5 miles}
& \textit{8.05 kilometres}
& Unit convention \\

Temperature
& \textit{68 degrees Fahrenheit}
& \textit{20 degrees Celsius}
& Unit convention \\

\bottomrule
\end{tabularx}

\caption{\small Illustrative linguistic and locale conventions used in the text benchmark audit. The examples show alternative valid forms; the detectors measure their occurrence in benchmark text rather than assigning correctness or user identity.}
\label{tab:example_contrasts}
\end{table*}


\subsection{Item Labels and Aggregation Metrics}
\label{sec:appendix_item_labeling}

For each item, we retain the detected convention signals
and assign one of four labels:
\begin{equation}
\begin{gathered}
\texttt{american\_only},\quad
\texttt{other\_only},\\
\texttt{mixed},\quad
\texttt{no\_match}.
\end{gathered}
\end{equation}

Items receive \texttt{american\_only} when they contain exclusively American-centered signals and \texttt{other\_only} when they contain exclusively comparison signals. Items containing both receive \texttt{mixed}; those with neither receive \texttt{no\_match}.

We summarize convention prevalence using
\emph{coverage}, the proportion of items containing
at least one audited signal:
\begin{equation}
\mathrm{Coverage}_b =
\frac{
\left|\left\{i\in b:
\ell(i)\neq\texttt{no\_match}\right\}\right|
}{
N_b
},
\label{eq:appendix_coverage}
\end{equation}
where \(N_b\) is the number of items in benchmark \(b\),
and \(\ell(i)\) is the assigned item label.

Coverage measures how frequently audited conventions appear in benchmark items. Reporting it alongside directional signal shares distinguishes the prevalence of identifiable conventions from the balance of forms among those matches.

Let \(A_b\) and \(O_b\) denote the counts of
American-centered and other user-valid signal matches
in benchmark \(b\), respectively. We define:
\begin{equation}
\mathrm{AmE\ Share}_b =
\frac{A_b}{A_b+O_b}.
\label{eq:appendix_ame_share}
\end{equation}

Mixed items may contribute to both \(A_b\) and \(O_b\). Coverage is therefore computed over items, whereas American-centered share is computed over matched signals.

For a signed representation of the same distribution,
we use:
\begin{equation}
\mathrm{Skew}_b =
2\,\mathrm{AmE\ Share}_b-1.
\label{eq:appendix_skew}
\end{equation}

Skew ranges from \(-1\) to \(1\): positive values indicate an American-centered majority, negative values indicate a majority of comparison signals, and zero denotes equal shares.

The comparison category includes several spelling traditions, date and time formats, and measurement conventions. The aggregate share measures the balance of this defined inventory rather than identifying a single alternative English variety.

\subsection{Convention-Level Distributions}
\label{sec:appendix_convention_distributions}

\begin{figure*}[t]
\centering
\includegraphics[width=1.0\linewidth]{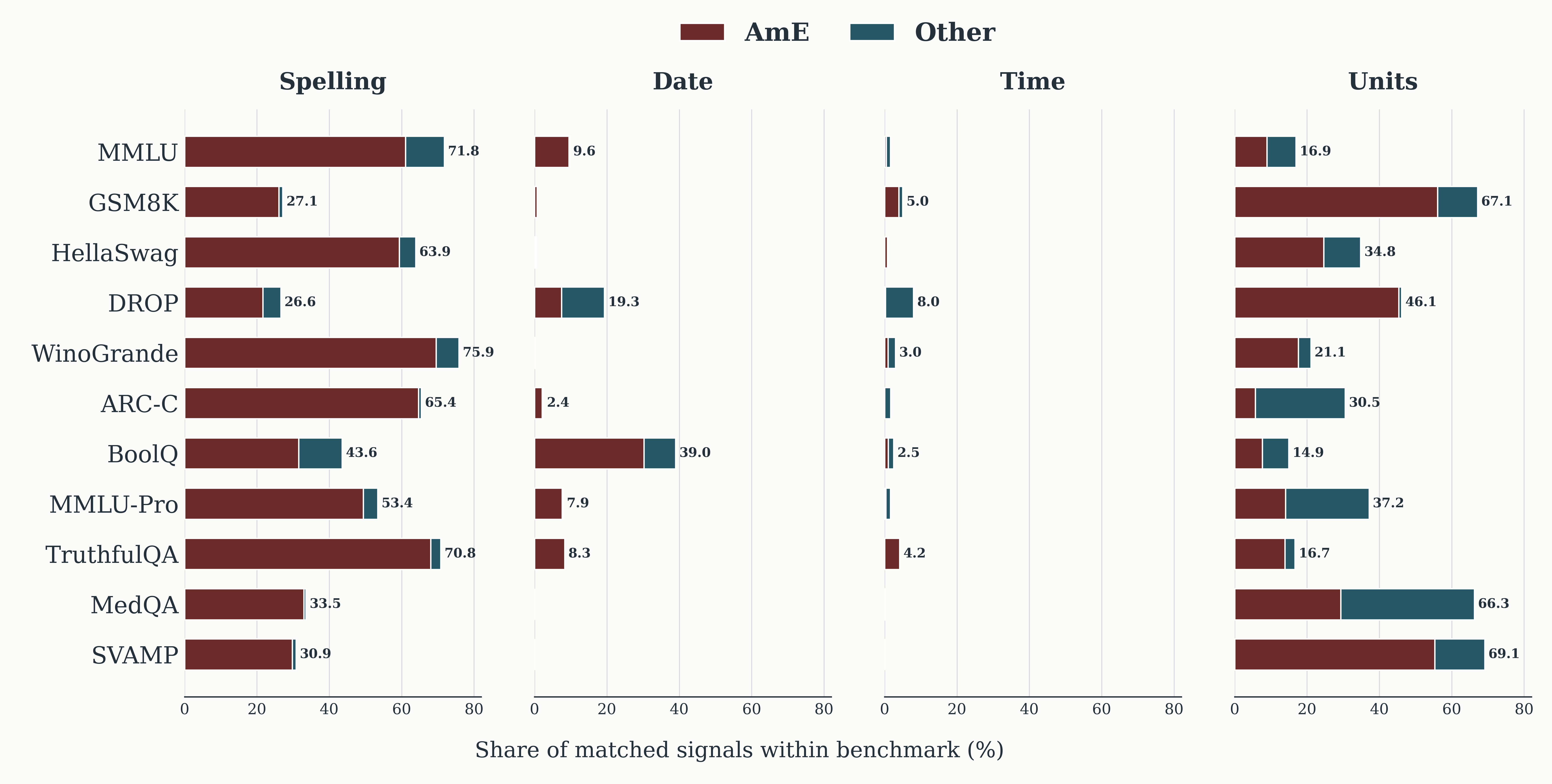}
\vspace{-6mm}
\caption{Benchmark-level distribution of linguistic and locale convention signals across 11 text benchmarks. Each panel shows, for a convention type, the proportions of American-centered and other user-valid matches within each benchmark. American-centered forms predominate in many benchmarks, with substantial variation in the direction of matches across spelling, dates, time, and measurement units. Date and time signals are generally sparser, revealing distinct convention profiles across resources.}

\label{fig:AmE_match_detail}
\end{figure*}

Main-paper Figure~\ref{fig:AmE_match_by_type} presents aggregate convention signals across benchmarks. Figure~\ref{fig:AmE_match_detail} separates spelling, dates, time, and measurement units, exposing the convention-specific patterns within each resource. Spelling contributes prominently to the American-centered profiles, with varying influence across benchmarks. Mathematics and reading-comprehension resources also contain date, time, and measurement signals, whereas some knowledge and commonsense resources are dominated by orthographic differences.

The category-level view separates each benchmark's aggregate direction from the convention classes contributing to it.

\subsection{Coverage and Spelling-Excluded Robustness}
\label{sec:appendix_coverage_robustness}

We report benchmark-level coverage and American-centered share separately to distinguish signal prevalence from directional concentration. Coverage spans 5.97\% in WinoGrande and 7.59\% in TruthfulQA to 65.85\% in MedQA and 78.08\% in DROP, demonstrating substantial differences in the proportion of items containing audited signals.

We repeat the directional analysis after removing all spelling matches. For benchmark \(b\), let \(A_b^{\neg S}\) and \(O_b^{\neg S}\) denote the remaining American-centered and other convention matches. We compute:

\begin{equation}
\mathrm{AmE\ Share}_b^{\neg S}
=
\frac{
A_b^{\neg S}
}{
A_b^{\neg S}+O_b^{\neg S}
}.
\label{eq:appendix_ame_no_spelling}
\end{equation}

Table~\ref{tab:rq1_coverage_robustness} shows an American-centered majority after removing spelling in eight of 11 benchmarks. TruthfulQA (90.5\%), GSM8K (83.3\%), SVAMP (80.0\%), and DROP (72.3\%) retain high shares. ARC-Challenge (22.5\%), MedQA (44.3\%), and MMLU-Pro (47.3\%) instead have majorities of other convention signals.

These results identify benchmarks whose aggregate profiles depend strongly on spelling and establish that the American-centered pattern also appears in nonspelling conventions.

\subsection{Scorer Case Study: Construction and Method}
\label{app:rq1_scoring_details}

We extend the convention audit from benchmark content to scoring behavior by testing how answer-text evaluators respond to meaning-preserving changes in linguistic or locale conventions.

\paragraph{Benchmark scoring procedures.} We study DROP~\citep{dua2019drop} using its official exact-match (EM) and token-level F1 implementation, including normalization rules, and TruthfulQA~\citep{lin2022truthfulqa} using the released generation evaluator's BLEU/ROUGE ranking of correct against incorrect reference answers.

These two evaluation interfaces enable controlled answer-text comparisons. Multiple-choice labels and extracted numerical answers use different scoring representations and are evaluated separately in their respective benchmark protocols.

\paragraph{Candidate construction and screening.} Each pair consists of an original reference-consistent answer and a meaning-preserving variant in regional spelling, written-date order, or time notation. We score both against identical gold annotations, isolating the effect of answer expression.

For DROP, we screen 23 answer-span candidates. Eleven originals do not receive full credit because they are incomplete gold answers. The resulting analysis contains 12 pairs from 11 questions: six spelling variants and six written-date variants.

For TruthfulQA, we screen 37 candidates from 27 questions. Three substitutions alter the brand name \textit{Flavor-Aid}, and one changes the official name \textit{World Health Organization}. Removing these four semantic changes yields 33 pairs from 25 questions: 28 spelling, three time-format, and two written-date variants.

These screening decisions preserve referenced entities and facts, ensuring that the paired test isolates differences in convention.

\paragraph{DROP scoring.} We apply DROP's official EM and F1 implementation to each original and variant. Originals are selected for full credit, yielding a common baseline. EM checks normalized answer agreement, while token-level F1 assigns partial credit for overlap.

\paragraph{TruthfulQA scoring.}
We reproduce the released generation evaluator's
BLEU/ROUGE comparison between correct and
incorrect references. Let \(R_q^+\) and \(R_q^-\)
denote the sets of correct and incorrect answers
for question \(q\), and let \(m(y,r)\) be the
similarity between candidate answer \(y\) and
reference \(r\) under metric \(m\).
The correctness indicator is:
\begin{equation}
\begin{aligned}
A_m(y,q)=\mathbf{1}\bigl\{&
\max_{r\in R_q^+}m(y,r)\\
&>\max_{r\in R_q^-}m(y,r)
\bigr\}.
\end{aligned}
\label{eq:truthfulqa_reference_ranking}
\end{equation}

The highest correct-reference similarity must strictly exceed the incorrect-reference similarity for a positive correctness decision. We preserve the released evaluator's correct reference ``I have no comment.'' when computing the ranking.

BLEU follows the released evaluator's sacreBLEU configuration and ROUGE uses ROUGE-score F1. For each original and variant, we compute the binary correctness indicator and the maximum similarity to a correct reference. The case study evaluates these released BLEU/ROUGE procedures; BLEURT and GPT-judge are separate evaluation interfaces.

\paragraph{Paired differences and uncertainty.} For every scoring metric, we calculate the variant-minus-original difference for each answer pair and report the mean paired change.

We estimate 95\% percentile confidence intervals with 10,000 question-clustered bootstrap replicates (seed 2026). Resampling questions keeps all variants from the same source question in one cluster, accounting for within-question dependence.

We report DROP EM and F1 separately and distinguish TruthfulQA binary correctness from continuous reference similarity. Benchmark-specific reporting preserves the different meanings and scales of the evaluation metrics.

\subsection{Scorer Case Study: Detailed Results}
\label{sec:appendix_scoring_results}

Table~\ref{tab:answer_surface_sensitivity-app} presents the complete paired results, including pair and question counts, original and variant scores, and question-clustered confidence intervals.

\begin{table*}[!tp]
\centering
\small
\renewcommand{\arraystretch}{1.16}
\setlength{\tabcolsep}{4.5pt}

\resizebox{\textwidth}{!}{%
\begin{tabular}{@{}llrrccc@{}}
\toprule
\textbf{Benchmark}
& \textbf{Scoring rule}
& \textbf{Pairs}
& \textbf{Questions}
& \textbf{Original}
& \textbf{Variant}
& \textbf{Change [95\% CI]} \\
\midrule

DROP
& Official EM
& 12 & 11
& 100.0\% & 33.3\%
& $-66.7$ [$-92.3,-36.4$] \\

DROP
& Official F1
& 12 & 11
& 100.0\% & 77.2\%
& $-22.8$ [$-41.6,-5.2$] \\

\midrule

TruthfulQA
& BLEU correctness
& 33 & 25
& 97.0\% & 97.0\%
& $0.0$ [$0.0,0.0$] \\

TruthfulQA
& ROUGE-1 correctness
& 33 & 25
& 97.0\% & 97.0\%
& $0.0$ [$0.0,0.0$] \\

TruthfulQA
& ROUGE-2 correctness
& 33 & 25
& 100.0\% & 93.9\%
& $-6.1$ [$-16.1,0.0$] \\

TruthfulQA
& ROUGE-Lsum correctness
& 33 & 25
& 100.0\% & 100.0\%
& $0.0$ [$0.0,0.0$] \\

\midrule

TruthfulQA
& BLEU correct-reference similarity
& 33 & 25
& 97.0 & 69.2
& $-27.7$ [$-34.0,-21.9$] \\

TruthfulQA
& ROUGE-1 correct-reference similarity
& 33 & 25
& 100.0 & 88.9
& $-11.1$ [$-13.5,-9.0$] \\

\bottomrule
\end{tabular}%
}

\caption{\small Detailed paired scorer sensitivity to meaning-preserving answer variants. DROP uses its official EM/F1 scorer; TruthfulQA reproduces the released BLEU/ROUGE correct-versus-incorrect reference-ranking procedure. EM, F1, and correctness changes are percentage points; reference similarities are reported on a 0--100 scale. Confidence intervals are calculated using 10,000 question-clustered bootstrap replicates. Benchmark results are not pooled.}
\label{tab:answer_surface_sensitivity-app}
\end{table*}

\paragraph{DROP: exact match and partial credit.} On 12 eligible pairs, official EM falls from 100.0\% to 33.3\%, a paired change of \(-66.7\) percentage points (95\% CI \([-92.3,-36.4]\)). Official F1 falls from 100.0\% to 77.2\%, a change of \(-22.8\) points (95\% CI \([-41.6,-5.2]\)).

Transformation type determines the amount of partial credit. The six spelling pairs lose an average of 45.7 F1 points (95\% CI \([-72.2,-23.0]\)); written-date variants retain full F1 even when EM decreases. Thus, token overlap recognizes valid alternatives that normalized exact match does not.

\paragraph{TruthfulQA: similarity and correctness.} Across 33 answer pairs, BLEU and ROUGE-1 correctness each remain at 97.0\%, and ROUGE-Lsum correctness remains at 100.0\%. ROUGE-2 correctness decreases from 100.0\% to 93.9\%, a 6.1-point change (95\% CI \([-16.1,0.0]\)).

Continuous reference similarity changes more substantially. Maximum correct-reference BLEU similarity falls from 97.0 to 69.2 (change \(-27.7\); 95\% CI \([-34.0,-21.9]\)), while maximum ROUGE-1 similarity falls from 100.0 to 88.9 (change \(-11.1\); 95\% CI \([-13.5,-9.0]\)). Both are measured on the reported 0--100 scale.

TruthfulQA's reference ranking can retain a correctness judgment as similarity to the closest correct reference decreases. This contrasts with the EM and partial-credit responses observed in DROP.

\paragraph{Supplementary exact-reference diagnostic.} As an additional diagnostic, we test normalized exact-reference matching on all 33 TruthfulQA variants. This strict matching rule rejects every transformed answer, illustrating sensitivity to answer expression. The diagnostic is separate from TruthfulQA's released generation metrics and is not reported as official TruthfulQA correctness.

\paragraph{Implications for evaluation normalization.} The paired results expose four distinct responses to meaning-preserving variation: loss of exact-match credit, partial token-overlap credit, reduced reference similarity, and unchanged reference-ranking correctness.

These findings motivate including meaning-preserving answer alternatives in scorer audits and documenting accepted spelling, date, time, and measurement conventions alongside evaluation rules. Such checks make recognition of equivalent responses measurable while retaining the underlying correctness criterion.

\FloatBarrier


\section{Geographic Visibility Audit}
\label{app:caption_geo_experiments}

This appendix documents the RQ2 audit of country visibility in five image--text resources, covering the data sources, geographic lexicon, country resolver, concentration analyses, socioeconomic comparisons, and WIT language-composition stress test (\cref{sec:rq2_geo_multimodal}). Geographic visibility throughout refers to explicit references in captions and associated textual metadata.

\subsection{Image--Text Resources}
\label{app:benchmarks}

We analyze DataComp1B, WIT English, RedCaps, CC12M, and COYO-700M, representing web-scale image--text data, Wikipedia-derived metadata, Reddit captions, web alt-text, and CommonCrawl-derived pairs. Table~\ref{tab:image-text-benchmarks} documents their characteristics and roles in the audit.

\begin{table*}[t]
\centering
\small
\renewcommand{\arraystretch}{1.45}
\setlength{\tabcolsep}{4.5pt}
\begin{tabularx}{\textwidth}{@{}p{0.15\textwidth} p{0.23\textwidth} p{0.20\textwidth} X@{}}
\toprule
\textbf{Benchmark} & \textbf{Source / formulation} & \textbf{Role in this study} & \textbf{Short description} \\ \hline
\midrule
\textbf{DataComp1B} \citep{gadre2023datacomp}
& Web-scale image--text data curation benchmark.
& Core web training-corpus benchmark.
& Broad, noisy web image--text source used to test whether country mentions in caption metadata reflect the geographic skew of large-scale multimodal training data. \\ \hline

\textbf{WIT English} \citep{srinivasan2021wit}
& Wikipedia image--text dataset with captions and structured page context.
& Curated encyclopedia-style control.
& English view of WIT, using caption and contextual text fields from Wikipedia. It provides a more structured source domain than web-crawl alt-text corpora. \\ \hline

\textbf{RedCaps} \citep{desai2021redcaps}
& Reddit-sourced image--text pairs from curated subreddits.
& Social-web comparison benchmark.
& User-generated social-media caption source. It tests whether geographic concentration persists outside Wikipedia and conventional web alt-text datasets. \\ \hline

\textbf{CC12M} \citep{changpinyo2021cc12m}
& Web-collected image--alt-text pairs extending Conceptual Captions.
& Classic alt-text caption benchmark.
& Large web caption corpus designed to broaden Conceptual Captions toward long-tail visual concepts. It gives a canonical vision--language pretraining comparison point. \\ \hline

\textbf{COYO-700M} \citep{byeon2022coyo700m}
& CommonCrawl-derived image--text pairs with additional metadata and filtering scores.
& Web-scale expansion benchmark.
& Open web-scale image--text corpus included to test whether the observed geographic patterns remain consistent beyond the smaller benchmark and control datasets. \\ \hline
\bottomrule
\end{tabularx}
\caption{Overview of the five image--text resources included in the geographic visibility audit. The table summarizes their data sources, characteristics, and roles in the analysis. Geographic visibility is measured using explicit country references resolved from captions and associated textual metadata. Country assignments reflect textual references rather than image geolocation or the geographic origin of the content.}
\label{tab:image-text-benchmarks}
\end{table*}

We analyze the textual component of each resource. A record is \emph{location-specified} when captions or associated metadata supply sufficient evidence for a country assignment; otherwise it is \emph{location-underspecified}.

Country visibility is calculated among location-specified records. The measure captures places referenced by captions and metadata, including contexts associated with described subjects, events, and entities. It measures geographic references in text rather than the geographic origin of photographs or contributors.

\subsection{Country Lexicon Construction}
\label{app:country_lexicon_construction}

Our geographic lexicon contains 3,413 entries: 3,036 country-name entries, 83 demonyms, and 294 city entries. Each record stores \texttt{country\_alpha2}, \texttt{country\_alpha3}, \texttt{canonical\_name}, \texttt{alias}, \texttt{alias\_norm}, \texttt{source}, \texttt{match\_type}, and \texttt{confidence\_tier}.

The lexicon combines four sources:

\begin{itemize}
    
    \item \textbf{Country names.} ISO-style records from \texttt{pycountry}\footnote{\url{https://github.com/pycountry/pycountry}}, including canonical and official names and their diacritic-folded forms.

    \item \textbf{Localized names.}
    Territory names from Babel/CLDR for English,
    French, German, Spanish, Portuguese, Japanese,
    Hindi, Greek, Simplified Chinese, and
    Traditional Chinese~\citep{unicodecldr2026}.

    \item \textbf{Aliases and demonyms.}
    Manually curated country abbreviations,
    alternative names, and adjectival references.

    \item \textbf{Cities and capitals.}
    National capitals from GeoNames through
    \texttt{geonamescache}\footnote{
    \url{https://github.com/yaph/geonamescache}},
    supplemented with selected high-signal
    cities~\citep{geonames2026}.
\end{itemize}

Country aliases include \texttt{USA}, \texttt{UK}, \texttt{UAE}, and \texttt{South Korea}; demonyms include \texttt{French}, \texttt{Japanese}, \texttt{Indian}, and \texttt{Brazilian}. The city layer covers Paris, London, Tokyo, New York, Mumbai, Sydney, and other selected locations.

We filter ambiguous city aliases, excluding digit-containing names, very short ASCII alternatives, common words, capital aliases shared across countries, and place names frequently used outside geographic contexts.

All aliases undergo Unicode NFKC normalization, lowercasing, punctuation and symbol normalization, and whitespace collapsing. Country names with diacritics retain original and ASCII-folded forms where available.

Each lexicon entry is typed as a country name, demonym, or city; that type determines its evidence weight in country assignment.

\subsection{Matching and Country Assignment}
\label{app:matching_and_country_assignment}

The resolver detects geographic mentions in captions and metadata and aggregates evidence for each candidate country.

ASCII aliases are matched as complete token sequences; non-ASCII aliases are matched in normalized Unicode text. Repeated matches of the same alias, country, and type within one record are deduplicated.

For each candidate country \(c\), we calculate:
\begin{equation}
s(c)=4n_{\mathrm{country}}(c)
+3n_{\mathrm{demonym}}(c)
+2n_{\mathrm{city}}(c),
\label{eq:appendix_country_score}
\end{equation}
where the three counts represent deduplicated
country-name, demonym, and city matches.

The weighting scheme prioritizes explicit country names over demonyms and city references. Cities provide useful geographic evidence but require more contextual discrimination.

Each record receives one of four outcomes:

\begin{itemize}
    \item \textbf{Single-country:}
    All matched geographic evidence identifies
    one country.

    \item \textbf{Weighted-majority:}
    Multiple countries are mentioned, but one
    receives a uniquely highest weighted score.

    \item \textbf{Ambiguous:}
    Multiple countries receive competing evidence
    without a unique highest score.

    \item \textbf{Unresolved:}
    No lexicon entry is matched.
\end{itemize}

Country frequencies include single-country and uniquely highest weighted-majority assignments. Ambiguous and unresolved records are excluded from the denominator, and ISO alpha-2 codes are mapped to canonical country names for reporting.

Country visibility describes the distribution of resolved textual assignments. Resolved coverage separately measures the proportion of records whose text supplies enough evidence for assignment.

\subsection{Geographic Concentration}
\label{app:top_bottom_country_analyses}

\begin{figure*}[t]
\centering
\includegraphics[width=\textwidth]{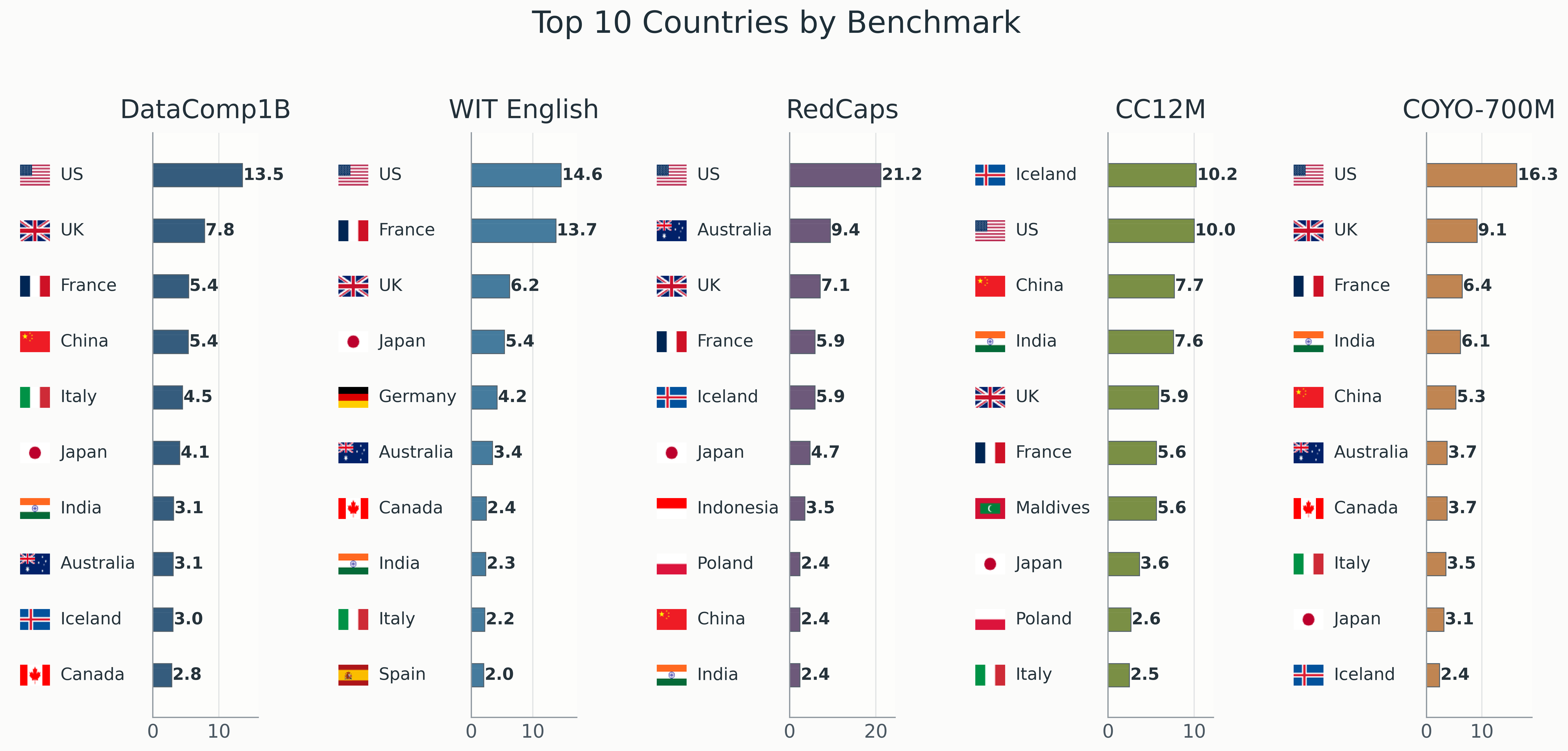}
\caption{\textbf{The high-frequency tail remains geographically concentrated.} Extending Figure~\ref{fig:top5_by_benchmark} to the top ten countries shows that resolved caption geography often centers on recurring countries in North America, Europe, and large Asian economies. CC12M is a benchmark-specific exception in which Iceland is the top resolved country, suggesting that source-pipeline artifacts or repeated content clusters may shape some country-frequency patterns. Rankings are benchmark-specific and do not require cross-benchmark overlap.}
\label{fig:top10_by_benchmark}
\end{figure*}

Main-paper Figure~\ref{fig:top5_by_benchmark} reports the five leading countries per resource; Figure~\ref{fig:top10_by_benchmark} extends these distributions to the top ten.

The United States leads in DataComp1B (13.55\%), WIT English (14.59\%), RedCaps (21.18\%), and COYO-700M (16.30\%) among resolved assignments.

CC12M exhibits a different ranking: Iceland (10.24\%), the United States (9.99\%), China (7.68\%), India (7.60\%), and the United Kingdom (5.86\%). This profile highlights variation across collection sources; repeated domains, alt-text conventions, or content clusters are potential contributors to the observed distribution.

Across the top-ten results, frequently referenced countries include the United States, United Kingdom, France, Australia, Canada, Japan, India, and China. These countries span North America, Europe, and major Asian economies.

Table~\ref{tab:bottom10-country} complements the leading-country view by listing the lowest-frequency countries with at least two resolved assignments in each resource.

\begin{table*}[t]
\centering
\scriptsize
\setlength{\tabcolsep}{3.5pt}
\renewcommand{\arraystretch}{1.12}
\resizebox{\textwidth}{!}{%
\begin{tabular}{rlllll}
\toprule
\textbf{Rank} & \textbf{DataComp1B} & \textbf{WIT English} & \textbf{RedCaps} & \textbf{CC12M} & \textbf{COYO-700M} \\
\midrule
1  & Belarus (2)    & Angola (2)                            & Anguilla (2)  & Afghanistan (2)          & Argentina (2) \\
2  & Belize (2)     & Bahrain (2)                           & Canada (2)    & Bahamas (2)              & Bangladesh (2) \\
3  & Bolivia (2)    & Belize (2)                            & China (2)     & Belize (2)               & Bulgaria (2) \\
4  & Congo (2)      & Benin (2)                             & Germany (2)   & Brunei Darussalam (2)    & Ethiopia (2) \\
5  & Dominica (2)   & Cameroon (2)                          & India (2)     & Denmark (2)              & Holy See (2) \\
6  & Guam (2)       & Dominica (2)                          & Italy (2)     & Falkland Islands (2)     & Jamaica (2) \\
7  & Lebanon (2)    & Guyana (2)                            & Maldives (2)  & Fiji (2)                 & Kuwait (2) \\
8  & Malta (2)      & Jamaica (2)                           & Poland (2)    & Guernsey (2)             & Moldova (2) \\
9  & Martinique (2) & Jordan (2)                            & Indonesia (3) & Iran (2)                 & Monaco (2) \\
10 & Monaco (2)     & Lao People's Democratic Republic (2)  & Japan (4)     & Jamaica (2)              & Nicaragua (2) \\
\bottomrule
\end{tabular}%
}
\caption{\textbf{Benchmark-wise low-frequency countries with at least two samples.} Each column lists the ten lowest-share countries within a benchmark, ranked independently over resolved country assignments. Countries are included only when they have at least two samples in that benchmark; counts in parentheses denote those caption-derived assignments. The resulting lists include small island territories, Caribbean and Pacific countries, and countries from Africa, Asia, Europe, and the Americas. This pattern suggests a broad long tail of sparsely represented geographies rather than a single region-specific effect.}
\label{tab:bottom10-country}
\end{table*}

The low-frequency lists span Africa, Asia, Europe, the Americas, the Caribbean, and the Pacific. Rankings are computed independently within each resource and reveal a geographically diverse tail of sparsely referenced countries. The contrast with the recurring leading countries illustrates concentration in the textual geography of these collections.

\subsection{GDP and Country Representation}
\label{app:gdp_correlation_experiment}

We link resolved country assignments to World Bank World Development Indicators through ISO identifiers and quantify their relationship with socioeconomic structure~\citep{worldbank2026wdi}.

For benchmark \(b\) and country \(c\), let
\(d_{c,b}\) denote the country's share of
resolved assignments and \(g_{c,b}\) its share
of GDP among the countries retained for that
benchmark. Our primary regression is:
\begin{equation}
\log_{10}d_{c,b}
=
\alpha_b+\beta_b\log_{10}g_{c,b}
+\epsilon_{c,b}.
\label{eq:appendix_gdp_model}
\end{equation}

We report log-space Pearson correlations, Spearman rank correlations, count-weighted correlations, and ordinary least-squares \(R^2\). Count weighting uses the number of resolved assignments for each country in correlations and regressions.

\paragraph{Socioeconomic indicators.}
We retrieve GDP in current U.S. dollars
(\texttt{NY.GDP.MKTP.CD}), GDP per capita
(\texttt{NY.GDP.PCAP.CD}), population
(\texttt{SP.POP.TOTL}), Internet usage
(\texttt{IT.NET.USER.ZS}), and international
tourism arrivals (\texttt{ST.INT.ARVL}).

GDP, GDP-per-capita, and population observations are predominantly from 2024; available tourism observations often refer to 2019 or 2020.

Country shares are normalized over aligned benchmark-specific sets with available required indicators. Regressions with Internet usage or tourism use their corresponding complete-case country sets.

Our primary specification retains countries with populations of at least one million. This threshold limits disproportionate effects from very small economies with unusually high tourism or web visibility relative to GDP and retains all five resources.

\begin{table*}[!tp]
\centering
\small
\renewcommand{\arraystretch}{1.12}
\setlength{\tabcolsep}{6pt}

\begin{tabular}{@{}lrrrrr@{}}
\toprule
\textbf{Benchmark}
& \textbf{Countries}
& \textbf{Pearson $r$}
& \textbf{Spearman $\rho$}
& \textbf{Weighted $r$}
& \textbf{OLS $R^2$} \\
\midrule
DataComp1B  & 118 & 0.816 & 0.786 & 0.907 & 0.667 \\
WIT English & 146 & 0.851 & 0.823 & 0.874 & 0.724 \\
RedCaps     & 30  & 0.647 & 0.735 & 0.809 & 0.419 \\
CC12M       & 74  & 0.770 & 0.772 & 0.833 & 0.593 \\
COYO-700M   & 109 & 0.817 & 0.775 & 0.897 & 0.667 \\
\midrule
\textbf{Mean}
& --
& \textbf{0.780}
& \textbf{0.778}
& \textbf{0.864}
& \textbf{0.614} \\
\bottomrule
\end{tabular}

\caption{GDP-share association with country
visibility across five image--text resources.
Countries with populations below one million are
excluded. Pearson correlations use log GDP share
and log dataset share; weighted correlations use
resolved country counts.}
\label{tab:gdp_recommended_setting}
\end{table*}

Table~\ref{tab:gdp_recommended_setting} shows positive GDP-share associations across all five resources. Spearman correlations range from 0.735 in RedCaps to 0.823 in WIT English. Mean OLS fit is \(R^2=0.614\), compared with count-weighted \(R^2=0.748\) reported in Table~\ref{tab:gdp_control_models}.

Countries with larger shares of economic activity generally receive greater visibility among the resolved textual country assignments.

\subsection{Socioeconomic Comparisons and Controls}
\label{app:gdp_control_variables}

We compare GDP share with population share, GDP per capita, Internet usage, and international tourism arrivals to characterize the socioeconomic associations in country visibility.

The response variable is:
\begin{equation}
y_{c,b}=\log_{10}d_{c,b},
\label{eq:appendix_geo_response}
\end{equation}
where \(d_{c,b}\) is the dataset share defined
in \cref{app:gdp_correlation_experiment}.

Single-predictor regressions use the logarithm of GDP share, population share, GDP per capita, Internet usage, or tourism arrivals. GDP and population shares are normalized within the retained resource-specific country set.

We additionally fit multivariable models that
combine GDP share with each control. For example,
let \(u_c\) denote the percentage of individuals
using the Internet in country \(c\). The
GDP-plus-Internet specification is:
\begin{equation}
\begin{aligned}
y_{c,b}={}&\alpha_b+\beta_1\log_{10}g_{c,b}\\
&+\beta_2\log_{10}u_c+\epsilon_{c,b}.
\end{aligned}
\label{eq:appendix_gdp_internet_model}
\end{equation}

The other multivariable specifications replace Internet usage with population share, GDP per capita, or tourism arrivals. Weighted regressions use resolved country counts as observation weights.

\begin{table}[!t]
\centering
\small
\renewcommand{\arraystretch}{1.12}
\setlength{\tabcolsep}{5pt}

\resizebox{\columnwidth}{!}{%
\begin{tabular}{@{}lrr@{}}
\toprule
\textbf{Model}
& \textbf{Mean OLS $R^2$}
& \textbf{Weighted $R^2$} \\
\midrule
GDP share only                    & 0.614 & 0.748 \\
Population share only             & 0.252 & 0.374 \\
GDP per capita only               & 0.210 & 0.254 \\
Internet users (\%) only          & 0.111 & 0.124 \\
Tourism arrivals only             & 0.347 & 0.396 \\
GDP share + population share      & 0.617 & 0.763 \\
GDP share + GDP per capita        & 0.617 & 0.763 \\
GDP share + Internet users (\%)   & 0.629 & 0.751 \\
GDP share + tourism arrivals      & 0.630 & 0.770 \\
\bottomrule
\end{tabular}%
}

\caption{Mean explanatory fit of socioeconomic
indicators across the five resources. Results follow
the population threshold and complete-case
requirements of the geographic analysis.}
\label{tab:gdp_control_models}
\end{table}

Table~\ref{tab:gdp_control_models} shows that GDP share has the strongest individual fit among the examined indicators. Tourism arrivals ranks second, followed by population share, GDP per capita, and Internet usage.

Adding a control to GDP share produces modest improvements: mean OLS \(R^2\) rises from 0.614 to 0.617--0.630 across the combined models.

These relationships are consistent with differences in economic scale and in the composition of web- and Wikipedia-derived collections. Source language and collection practices also influence the references available for geographic resolution.

\subsection{Language-Composition Stress Test on WIT}
\label{app:wit_language_stress_test}

We use WIT to examine geographic visibility under a different language composition, constructing balanced samples in French, German, Spanish, and Japanese. We apply the country-resolution procedure from \cref{app:matching_and_country_assignment} to each language setting. Country shares are computed from records whose caption or associated metadata yields a resolved country assignment.

\paragraph{Cross-language geographic matching.} The shared resolver detects country names, aliases, demonyms, and selected city references in multilingual text. Place names such as \textit{France}, \textit{Paris}, \textit{Berlin}, \textit{Canada}, \textit{Mexico}, \textit{Chile}, and \textit{Argentina} appear in similar forms across several Latin-script languages. Japanese records may also contain romanized place names, URLs, titles, entity names, and English aliases. Applying the same resolver keeps the detection inventory consistent across samples, enabling controlled comparison of resolved geographic references.

\paragraph{Geographic visibility across languages.} Table~\ref{tab:wit_language_stress_test} presents the language-composition comparison. The U.S. share falls from 20.0\% in the corresponding English comparison to 8.5\% in the balanced non-English sample, demonstrating that language composition materially changes the observed country distribution.

\begin{table}[t]
\centering
\small
\renewcommand{\arraystretch}{1.15}
\setlength{\tabcolsep}{4pt}

\begin{tabular*}{\columnwidth}{
@{\extracolsep{\fill}}lcc@{}}
\toprule
\textbf{Measure}
& \textbf{English}
& \shortstack{\textbf{Balanced}\\\textbf{non-English}} \\
\midrule
U.S. share (\%)
& 20.0 & 8.5 \\
Top-10 GDP share (\%)
& -- & 53.5 \\
Pearson $r$
& -- & 0.752 \\
Spearman $\rho$
& -- & 0.744 \\
\bottomrule
\end{tabular*}

\caption{Geographic visibility in the WIT
language-composition stress test. The balanced
non-English sample includes French, German,
Spanish, and Japanese. Country shares use
resolved assignments; GDP correlations and
top-ten concentration are reported for the
non-English setting.}

\label{tab:wit_language_stress_test}
\end{table}

Economic concentration remains substantial: the ten countries with the largest GDP shares account for 53.5\% of resolved assignments in the non-English sample, with Pearson \(r=0.752\) and Spearman \(\rho=0.744\). The comparison identifies two concurrent patterns: changing the language composition shifts the visible countries, while economically prominent countries continue to account for a substantial share of resolved geographic references.

\paragraph{Interpretation.} Prior work documents concentrated language and source composition in English web corpora such as C4~\citep{dodge-etal-2021-documenting}. Our WIT analysis extends this concern to textual geography in image--text collections, showing that the language of captions and metadata is associated with which countries become visible.

The main WIT English audit and the language-comparison experiment have distinct analysis settings. The U.S. shares of 14.59\% in the main audit and 20.0\% in the separate English comparison therefore refer to different samples and are reported independently. Reporting geographic coverage alongside language composition and collection sources makes these influences inspectable and supports more precise interpretation of cross-resource country distributions.


\section{Documentation Audit}
\label{sec:documentation_audit_agreement}

This appendix documents the RQ3 documentation audit, including the annotation protocol, inter-annotator agreement, and resource-level results (\cref{sec:documentation_audit}). The central measure is whether public documentation states user-relevant assumptions and makes them inspectable.

\paragraph{Annotation procedure.}
Two computer science doctoral students, one author and one external volunteer, independently reviewed public documentation for all 16 resources, including papers, official dataset cards, repositories, and the sources listed in Table~\ref{tab:documentation_audit_sources}. Both annotators had experience in NLP evaluation, benchmark analysis, and dataset documentation. Both voluntarily agreed to participate and consented to sharing their annotation data. Before annotation, both reviewers established a shared interpretation of the six criteria and their three-level scoring rubric:

\begin{itemize}
    \item \textbf{0 (Absent):} The relevant information
    is not documented.

    \item \textbf{1 (Mentioned):} The documentation
    explicitly describes the relevant context or
    assumption.

    \item \textbf{2 (Operationalized):} The information
    is available through an inspectable metadata field,
    source partition, annotation protocol, scoring rule,
    or comparable mechanism.
\end{itemize}

All nonzero scores require explicit documentary evidence. Where several sources exist for a resource, we retain the highest evidence-supported score. Operationalization requires an inspectable field, protocol, rule, or other mechanism that a benchmark user can apply.

\paragraph{Source and geographic metadata.} This criterion covers website URLs, Internet domains, top-level domains (TLDs), language tags, and country or region fields that support provenance tracing and geographic auditing. A \emph{domain} here is an Internet website domain, such as \texttt{bbc.co.uk}; its \texttt{.uk} country-code TLD is associated with the United Kingdom. Likewise, \texttt{.ca} and \texttt{.jp} indicate country-code associations with Canada and Japan, whereas a generic TLD such as \texttt{.com} does not specify a country. Such signals can help trace sources while remaining distinct from content creation or contributor locations.

\paragraph{Inter-annotator agreement.} Table~\ref{tab:documentation_audit_agreement} gives agreement before adjudication: exact agreement is 99.0\% across criterion-level decisions, with overall Cohen's \(\kappa=0.984\).

\begin{table}[t]
\centering
\small
\renewcommand{\arraystretch}{1.14}
\setlength{\tabcolsep}{3pt}

\begin{tabularx}{\columnwidth}{
    >{\raggedright\arraybackslash}X
    >{\centering\arraybackslash}p{0.13\columnwidth}
    >{\centering\arraybackslash}p{0.13\columnwidth}
    >{\centering\arraybackslash}p{0.13\columnwidth}
}
\toprule
\rowcolor{headergray}
\textbf{Criterion}
& \makecell{\textbf{Exact}\\\textbf{(\%)}}
& \textbf{$\kappa$}
& \textbf{Disagr.} \\
\midrule

Source-user context        & 100.0 & --    & 0 \\
Target/eval. context       & 100.0 & --    & 0 \\
Annotator visibility       & 100.0 & 1.000 & 0 \\
Locale assumptions         & 100.0 & 1.000 & 0 \\
Evaluation normalization   & 100.0 & 1.000 & 0 \\
Source/geographic metadata & 93.8  & 0.881 & 1 \\

\midrule
\textbf{Overall (micro)}
& \textbf{99.0}
& \textbf{0.984}
& \textbf{1} \\
\bottomrule
\end{tabularx}

\caption{Inter-annotator agreement across 16 resources before adjudication. Exact agreement and Cohen's $\kappa$ are reported for each criterion. Dashes indicate undefined $\kappa$ when both annotators assign a constant score.}
\label{tab:documentation_audit_agreement}
\end{table}

The single disagreement concerns CC12M's released image URLs. One annotator judged these sufficient to operationalize source/geographic metadata; the other assigned a lower score. Adjudication assigned score 2 because the released URLs permit direct inspection of source provenance. The high agreement reflects the rubric's evidence-based criteria and supports consistent interpretation of documentation availability across resources.

\paragraph{Resource-level documentation results.} Table~\ref{tab:documentation_audit_sources} identifies the consulted documentation, and Table~\ref{tab:documentation_audit_benchmarkwise_scores} reports all six criterion scores for each resource.

\begin{table*}[t]
\centering
\footnotesize
\renewcommand{\arraystretch}{1.15}
\setlength{\tabcolsep}{3.5pt}
\begin{tabularx}{\textwidth}{@{}p{0.15\textwidth} p{0.13\textwidth} X@{}}
\toprule
\textbf{Resource} & \textbf{Family} & \textbf{Primary documentation source} \\
\midrule
MMLU 
& Text QA 
& \citet{hendrycks2021mmlu}, \href{https://openreview.net/forum?id=d7KBjmI3GmQ}{paper}. \\

GSM8K 
& Text QA 
& \citet{cobbe2021gsm8k}, \href{https://arxiv.org/abs/2110.14168}{paper and dataset release}. \\

HellaSwag 
& Text QA 
& \citet{zellers2019hellaswag}, \href{https://aclanthology.org/P19-1472/}{paper and supplement}. \\

DROP 
& Text QA 
& \citet{dua2019drop}, \href{https://aclanthology.org/N19-1246/}{paper}. \\

WinoGrande 
& Text QA 
& \citet{sakaguchi2020winogrande}, \href{https://doi.org/10.1145/3474381}{paper and project page}. \\

ARC-Challenge 
& Text QA 
& \citet{clark2018arc}, \href{https://arxiv.org/abs/1803.05457}{paper}. \\

BoolQ 
& Text QA 
& \citet{clark2019boolq}, \href{https://aclanthology.org/N19-1300/}{paper}. \\

MMLU-Pro 
& Text QA 
& \citet{wang2024mmlupro}, \href{https://openreview.net/forum?id=y10DM6R2r3}{paper and supplement}. \\

TruthfulQA 
& Text QA 
& \citet{lin2022truthfulqa}, \href{https://aclanthology.org/2022.acl-long.229/}{paper}. \\

MedQA 
& Text QA 
& \citet{jin2021medqa}, \href{https://www.mdpi.com/2076-3417/11/14/6421}{paper}. \\

SVAMP 
& Text QA 
& \citet{patel2021svamp}, \href{https://aclanthology.org/2021.naacl-main.168/}{paper}. \\
\midrule
DataComp1B 
& Image--text 
& \citet{gadre2023datacomp}, \href{https://openreview.net/forum?id=dVaWCDMBof}{DataComp paper and dataset card}. \\

WIT English 
& Image--text 
& \citet{srinivasan2021wit}, \href{https://github.com/google-research-datasets/wit}{paper and dataset repository}. \\

RedCaps 
& Image--text 
& \citet{desai2021redcaps}, \href{https://huggingface.co/datasets/kdexd/red_caps}{paper and dataset card}. \\

CC12M 
& Image--text 
& \citet{changpinyo2021cc12m}, \href{https://github.com/google-research-datasets/conceptual-12m}{Conceptual 12M paper and official repository}. \\

COYO-700M 
& Image--text 
& \citet{byeon2022coyo700m}, \href{https://huggingface.co/datasets/kakaobrain/coyo-700m}{dataset card and repository}. \\
\bottomrule
\end{tabularx}
\caption{Resources included in the documentation audit and the public documentation sources used for scoring. Each link points to the benchmark paper, official dataset card, or official repository consulted during the audit. All documentation sources were accessed in April 2026.}
\label{tab:documentation_audit_sources}
\end{table*}

\begin{table*}[!t]
\centering
\footnotesize
\renewcommand{\arraystretch}{1.15}
\setlength{\tabcolsep}{4pt}

\resizebox{0.82\textwidth}{!}{%
\begin{tabular}{@{}llrrrrrrr@{}}
\toprule
\textbf{Resource}
& \textbf{Family}
& \textbf{Source}
& \textbf{Target/eval.}
& \textbf{Annot.}
& \textbf{Locale}
& \textbf{Eval.}
& \textbf{Src./Geo.}
& \textbf{Total} \\
\midrule
MMLU          & Text QA    & 2 & 1 & 0 & 0 & 2 & 0 & 5/12 \\
GSM8K         & Text QA    & 2 & 1 & 1 & 0 & 2 & 0 & 6/12 \\
HellaSwag     & Text QA    & 2 & 1 & 1 & 0 & 2 & 0 & 6/12 \\
DROP          & Text QA    & 2 & 1 & 1 & 0 & 2 & 0 & 6/12 \\
WinoGrande    & Text QA    & 2 & 1 & 1 & 0 & 2 & 0 & 6/12 \\
ARC-Challenge & Text QA    & 2 & 1 & 0 & 0 & 2 & 0 & 5/12 \\
BoolQ         & Text QA    & 2 & 1 & 1 & 0 & 2 & 0 & 6/12 \\
MMLU-Pro      & Text QA    & 2 & 1 & 1 & 0 & 2 & 0 & 6/12 \\
TruthfulQA    & Text QA    & 2 & 1 & 1 & 0 & 2 & 0 & 6/12 \\
MedQA         & Text QA    & 2 & 1 & 0 & 1 & 2 & 1 & 7/12 \\
SVAMP         & Text QA    & 2 & 1 & 0 & 0 & 2 & 0 & 5/12 \\
\midrule
DataComp1B    & Image--text & 2 & 1 & 0 & 0 & 1 & 2 & 6/12 \\
WIT English   & Image--text & 2 & 1 & 0 & 2 & 0 & 2 & 7/12 \\
RedCaps       & Image--text & 2 & 1 & 1 & 1 & 0 & 2 & 7/12 \\
CC12M         & Image--text & 2 & 1 & 0 & 0 & 0 & 2 & 5/12 \\
COYO-700M     & Image--text & 2 & 1 & 1 & 1 & 0 & 2 & 7/12 \\
\bottomrule
\end{tabular}%
}

\caption{Benchmark-wise documentation audit scores across
16 resources. Each criterion is scored as 0 (absent),
1 (mentioned), or 2 (operationalized), with a maximum
total of 12. Source = source-user context;
Target/eval. = target-user or evaluation context;
Annot. = annotator visibility; Locale = locale assumptions;
Eval. = evaluation normalization; Src./Geo. =
source/geographic metadata.}
\label{tab:documentation_audit_benchmarkwise_scores}
\end{table*}

The resource-level scores show a systematic source--target gap. All 16 resources operationalize source-user context through documented collection sources such as examinations, Wikipedia, Reddit, and web content. All mention target-user or evaluation context, but none operationalize the link between evaluation scores and intended user populations or settings.

Annotator visibility and locale assumptions are less consistently documented. Evaluation normalization is comparatively well specified, particularly for text benchmarks, although the accepted alternatives and scoring behaviors vary across evaluation procedures. These results establish that source provenance and the intended scope of evaluation claims are documented at different levels of specificity. Reporting both makes benchmark interpretation more transparent and supports the user-centric checklist in \cref{sec:discussion}.


\section{Example User-Centric Documentation Template}
\label{app:user_centric_template}

The checklist in Table~\ref{tab:user_centric_documentation_checklist}
can be implemented through the following reporting fields.
For each field, benchmark creators should document the
available evidence, relevant assumptions, and information
needed to interpret evaluation results.

\begin{enumerate}
    \item \textbf{Source users.}
    Describe who produced the benchmark content and through
    which platforms, institutions, tasks, or collection
    processes. Indicate whether source populations are
    directly observed, inferred, or unknown.

    \item \textbf{Target users and evaluation context.}
    Specify the users, settings, and evaluative claims
    the benchmark is intended to represent. Identify
    contexts where its applicability remains uncertain,
    including different linguistic, regional, or
    domain-specific settings.

    \item \textbf{Annotators and label provenance.}
    Describe who produced labels, reference answers,
    or quality judgments, including available information
    about expertise, locale, recruitment, compensation,
    and quality control. For automatically derived labels,
    report their source and validation procedure.

    \item \textbf{Locale conventions.}
    Document spelling, date, time, measurement, currency,
    numeric-format, language-variety, and source-language
    conventions in benchmark content and scoring.
    Distinguish conventions inherited from source data
    from those introduced during benchmark construction.

    \item \textbf{Evaluation normalization.}
    Specify how scoring procedures handle valid
    alternative answers, including differences in
    spelling, formatting, time, dates, and measurement
    conventions. Document which forms are accepted,
    normalized, converted, or rejected, and whether
    semantically equivalent responses receive
    consistent scores.

    \item \textbf{Source and geographic metadata.}
    Report available source URLs, website domains,
    top-level domains, language tags, country or region
    fields, and related provenance information. Clarify
    whether these fields identify a content source,
    referenced location, depicted entity, or
    another geographic signal.

    \item \textbf{Known coverage gaps.}
    Identify gaps in user populations, linguistic
    conventions, geographic visibility, task settings,
    or annotation provenance that affect the
    interpretation of benchmark results.

    \item \textbf{Recommended and non-recommended uses.}
    State which evaluation tasks and comparisons the
    benchmark supports, and which broader claims
    require additional evidence, particularly for
    multilingual, cross-regional, or deployment settings.

    \item \textbf{Auditability and updates.}
    Identify the released metadata, documentation,
    scoring code, and other resources needed to
    reproduce or extend the audit. Describe how changes
    to benchmark content, scoring rules, or documentation
    are recorded.
\end{enumerate}


\section{AI Use Statement}
\label{sec:AI-usage}

Generative AI tools were used in limited assistive roles for \textbf{(1) text editing}, including grammar, spelling, phrasing, and clarity, and \textbf{(2) code assistance}, including routine implementation, data preprocessing, filtering, and debugging. 

All AI-assisted text and code were reviewed and refined by the authors, and AI-assisted code was manually verified and tested before use. The authors independently formulated the research questions, developed the methodology, designed and conducted the experiments, analyzed and interpreted the results, and determined the final conclusions. The authors take full responsibility for the content of this work.


\end{document}